\documentclass[conference]{IEEEtran}
\IEEEoverridecommandlockouts
\usepackage{cite}
\usepackage{amsmath,amssymb,amsfonts}
\usepackage{algorithmic}
\usepackage{graphicx}
\usepackage{textcomp}
\usepackage{xcolor}
\usepackage{booktabs}
\usepackage{multirow}
\usepackage{makecell}
\usepackage{subfigure}
\usepackage{hyperref}
\usepackage{fontawesome}
\usepackage[linesnumbered,ruled,vlined]{algorithm2e}
\usepackage{fancyhdr}

\usepackage{bm}
\newtheorem{definition}{\bf Definition}
\def\BibTeX{{\rm B\kern-.05em{\sc i\kern-.025em b}\kern-.08em
    T\kern-.1667em\lower.7ex\hbox{E}\kern-.125emX}}
\hypersetup{
colorlinks=true,
linkcolor=black,
citecolor=black,
urlcolor=darkgray
}

\begin{document}

\title{Collaborative Pareto Set Learning in Multiple Multi-Objective Optimization Problems\\
\thanks{This work was supported in part by the National Natural Science Foundation of China (62006044) and in part by the Programme of Science and Technology of Guangdong Province (202201010377).}
}

\DeclareRobustCommand*{\IEEEauthorrefmark}[1]{%
    \raisebox{0pt}[0pt][0pt]{\textsuperscript{\footnotesize\ensuremath{#1}}}}
\author{
\IEEEauthorblockN{Chikai Shang\IEEEauthorrefmark{1}, Rongguang Ye\IEEEauthorrefmark{2}, Jiaqi Jiang\IEEEauthorrefmark{1}, Fangqing Gu\IEEEauthorrefmark{1}$^\ast$}
\IEEEauthorblockA{
\IEEEauthorrefmark{1} \textit{School of Mathematics and Statistics, Guangdong University of Technology, China}
}
\IEEEauthorblockA{
\IEEEauthorrefmark{2} \textit{Department of Computer Science and Engineering, Southern University of Science and Technology, China}
}
\IEEEauthorblockA{3121006952@mail2.gdut.edu.cn,~yerg2023@mail.sustech.edu.cn,~3221007074@mail2.gdut.edu.cn,~fqgu@gdut.edu.cn}
}

\maketitle

\begin{abstract}
Pareto Set Learning (PSL) is an emerging research area in multi-objective optimization, focusing on training neural networks to learn the mapping from preference vectors to Pareto optimal solutions. However, existing PSL methods are limited to addressing a single Multi-objective Optimization Problem (MOP) at a time. When faced with multiple MOPs, this limitation results in significant inefficiencies and hinders the ability to exploit potential synergies across varying MOPs. In this paper, we propose a Collaborative Pareto Set Learning (CoPSL) framework, which learns the Pareto sets of multiple MOPs simultaneously in a collaborative manner. CoPSL particularly employs an architecture consisting of shared and MOP-specific layers. The shared layers are designed to capture commonalities among MOPs collaboratively, while the MOP-specific layers tailor these general insights to generate solution sets for individual MOPs. This collaborative approach enables CoPSL to efficiently learn the Pareto sets of multiple MOPs in a single execution while leveraging the potential relationships among various MOPs. To further understand these relationships, we experimentally demonstrate that shareable representations exist among MOPs. Leveraging these shared representations effectively improves the capability to approximate Pareto sets. Extensive experiments underscore the superior efficiency and robustness of CoPSL in approximating Pareto sets compared to state-of-the-art approaches on a variety of synthetic and real-world MOPs. Code is available at~\href{https://github.com/ckshang/CoPSL}{\faGithub~\texttt{https://github.com/ckshang/CoPSL}}.

\end{abstract}


\section{Introduction}
Multi-objective Optimization Problems (MOPs) frequently occur in many real-world applications, such as drug discovery~\cite{dara2022machine}, software engineering~\cite{harman2010software}, etc. In MOPs, the objectives are typically in conflict, rendering it impossible to optimize all of them simultaneously with a single solution~\cite{gong2023effects}. Instead, there are Pareto optimal solutions, each representing a different trade-off among the objectives. The primary goal in solving MOPs is to find a set of these Pareto optimal solutions, known as the Pareto set. The image of the Pareto set in the objective space is termed the Pareto front.

Over the past several decades, researchers have developed various population-based Evolutionary Multi-objective Optimization (EMO) algorithms~\cite{deb2002fast, zhang2007moea, deb2013evolutionary}. These algorithms are designed to optimize a population of solutions, ultimately obtaining a well-distributed set of solutions along the Pareto front. However, a key limitation of these population-based approaches is that merely a finite set of approximate Pareto solutions can be obtained~\cite{lin2022pareto}. In response to this limitation, a novel model-based approach known as Pareto Set Learning (PSL) has recently been proposed~\cite{lin2022pareto, lin2023evolutionary, ruchte2021scalable}. PSL focuses on training a neural network model to approximate the entire Pareto set of a given MOP. In this approach, preference vectors serve as inputs to the model, which maps them to Pareto solutions. These solutions are then used to calculate their objective values, which are subsequently employed to refine the model via an optimization process that utilizes a loss function, specifically the aggregation functions used in decomposition-based EMO~\cite{zhang2007moea}.

However, most current PSL approaches are designed to handle merely a single MOP. When confronted with multiple MOPs, these approaches typically learn the Pareto set of each MOP sequentially and independently, resulting in significant inefficiencies in both time and memory. Furthermore, it is worth noting that there exist latent synergies between different MOPs~\cite{bali2019multifactorial}. Such sequential and independent learning approaches fail to capitalize on the potential common relationships across these problems effectively. This neglect is especially notable in model-based approaches, where there is a high likelihood of shareable common representations across various MOPs.

In this paper, we propose a \textbf{Co}llaborative \textbf{P}areto \textbf{S}et \textbf{L}earning (CoPSL) framework, which aims to solve multiple MOPs simultaneously in a collaborative fashion. Drawing inspiration from hard parameter sharing in multi-task learning~\cite{vandenhende2021multi}, our CoPSL architecture is divided into layers that are either shared across MOPs or dedicated to specific MOPs. This architecture enables shared layers to collaboratively learn common relationships across MOPs, which are then utilized by MOP-specific layers to generate the Pareto set for each MOPs. It allows CoPSL to optimize multiple MOPs simultaneously with greater efficiency in both time and memory.

To further explore the relationships among different MOPs, we delve into the working mechanics of CoPSL through experimental investigations, demonstrating the existence of common representations that can be shared among MOPs. Leveraging these collaboratively learned common representations helps the framework better approximate Pareto sets. To evaluate the efficacy of CoPSL, we compare it with widely used EMO algorithms and state-of-the-art PSL approaches. The experimental results demonstrate that the proposed CoPSL operates with remarkable efficiency and outperforms the comparative algorithms in terms of Pareto set approximation, albeit by a modest yet consistent margin.

In summary, our key contributions are as follows:
\begin{itemize}
    \item We propose a collaborative Pareto set learning framework that collaboratively learns common relationships across multiple MOPs through shared layers, which are then utilized by MOP-specific layers to map them to solutions. This collaborative learning enables more efficient handling of multiple MOPs in a single run.
    \item We study the working principle of CoPSL through experimental investigations. It reveals that shareable representations exist within CoPSL, which helps the framework better approximate the Pareto set of each MOP.
\end{itemize}

The remainder of this paper is organized as follows. In Section~\ref{section2}, we introduce the preliminary concepts. Section~\ref{section3} and Section~\ref{section4} provide a detailed description and further analysis of CoPSL. Section~\ref{section5} presents experimental studies comparing CoPSL with multiple state-of-the-art algorithms. Finally, the conclusion and future work are discussed in Section~\ref{section6}.

\section{Preliminaries} \label{section2}

\subsection{Multi-objective Optimization}

Without loss of generality, a multi-objective optimization problem with $m$ ($m\geq 2$) objectives and $n$ decision variables can be formulated as follows:
\begin{equation}
    \min_{\mathbf{x}\in\mathcal{X}}~ \mathbf{F}(\mathbf{x}) = \big(f_1(\mathbf{x}), \ldots, f_m(\mathbf{x})\big)^T
\end{equation}
where $\mathbf{x}$ is a solution in the decision space $\mathcal{X}\subset\mathbb{R}^n$, and $\mathbf{F}:\mathcal{X} \rightarrow \mathbb{R}^m$ is an objective function vector. For an MOP, no single solution can optimize all objectives at the same time, and we have to make a trade-off among them. We have the following definitions of multi-objective optimization: 

\begin{definition}[Pareto Dominance]
Let $\mathbf{x},\mathbf{y}\in\mathcal{X}$, $\mathbf{x}$ is said to dominate $\mathbf{y}$ (denoted as $\mathbf{x} \prec \mathbf{y}$) if and only if $\forall i \in [m], f_i(\mathbf{x})\leq f_i(\mathbf{y})$ and $\exists j\in [m], f_j(\mathbf{x})<f_j(\mathbf{y})$.
\end{definition}

\begin{definition}[Pareto Optimality]
A solution $\mathbf{x}^\ast\in\mathcal{X}$ is said to be Pareto optimal if there is no other solution $\mathbf{x}\in\mathcal{X}$ such that $\mathbf{x}\prec\mathbf{x}^\ast$.
\end{definition}

\begin{definition}[Pareto Set/Front]
The set consisting of all Pareto optimal solutions is called the Pareto set: $PS=\{\mathbf{x}\in\mathcal{X}\mid \forall \mathbf{y} \in \mathcal{X}, \mathbf{y} \nprec \mathbf{x} \}$ and the corresponding objective vector set of the PS is the Pareto front: $PF = \{ \mathbf{F}(\mathbf{x})\mid\mathbf{x}\in PS \}$.
\end{definition}

In general, the goal of optimizing an MOP is to approximate both the Pareto set and the Pareto front.

\subsection{Pareto Set Learning}

Pareto set learning utilizes a neural network on user-defined preferences to approximate the entire Pareto set~\cite{lin2022pareto, lin2023evolutionary, lin2022pareto2}. Specifically, a preference vector $\mathbf{p}=(p_1, \ldots, p_m) \in \mathbb{R}^m_+$ (with $\sum_{i=1}^mp_i=1$) indicates the desired trade-off among objectives. The network takes the preference vectors as inputs, with the network outputting tailored solutions. Typically, an aggregation function is employed as the loss function, enabling the optimization of MOPs through gradient-based training~\cite{kingma2014adam}. Such an optimization process ensures that the generated solutions are high quality and align closely with the defined preferences. This approach has been successfully applied across various domains, including deep multi-task learning~\cite{navon2021learning, hoang2023improving}, neural multi-objective combinatorial optimization~\cite{lin2022pareto2}, and so on.

\subsection{Multi-task Learning}

Multi-task learning (MTL) leverages shared representations to facilitate collaborative learning from multiple tasks. MTL architectures are generally categorized into hard or soft parameter sharing techniques~\cite{vandenhende2021multi}. Hard parameter sharing involves a shared encoder that branches out into task-specific decoding heads~\cite{kokkinos2017ubernet, chen2018gradnorm}. In contrast, soft parameter sharing assigns separate parameter sets to each task and utilizes feature sharing mechanisms to enable cross-task communication~\cite{misra2016cross}. Due to the shared layer and information sharing, hard parameter sharing methods often outperform their single-task counterparts in both resource utilization efficiency and per-task performance.

\section{Collaborative Pareto Set Learning} \label{section3}
In the CoPSL framework, we simultaneously tackle $K$ MOPs, each characterized by the same $m$ dimensions. This setup can be described as follows:
\begin{equation}
    \min_{\mathbf{x}_i\in\mathcal{X}_i,i\in[K]}~
    \begin{cases} 
        \mathbf{F}_1(\mathbf{x}_1) = \big(f^1_1(\mathbf{x}_1), \ldots, f^1_m(\mathbf{x}_1)\big)^T \\
        \mathbf{F}_2(\mathbf{x}_2) = \big(f^2_1(\mathbf{x}_2), \ldots, f^2_m(\mathbf{x}_2)\big)^T \\
        \cdots \\
        \mathbf{F}_K(\mathbf{x}_K) = \big(f^K_1(\mathbf{x}_K), \ldots, f^K_m(\mathbf{x}_K)\big)^T 
    \end{cases}
\end{equation}

\begin{figure*}[!t]
    \centering
    \includegraphics[scale=0.95]{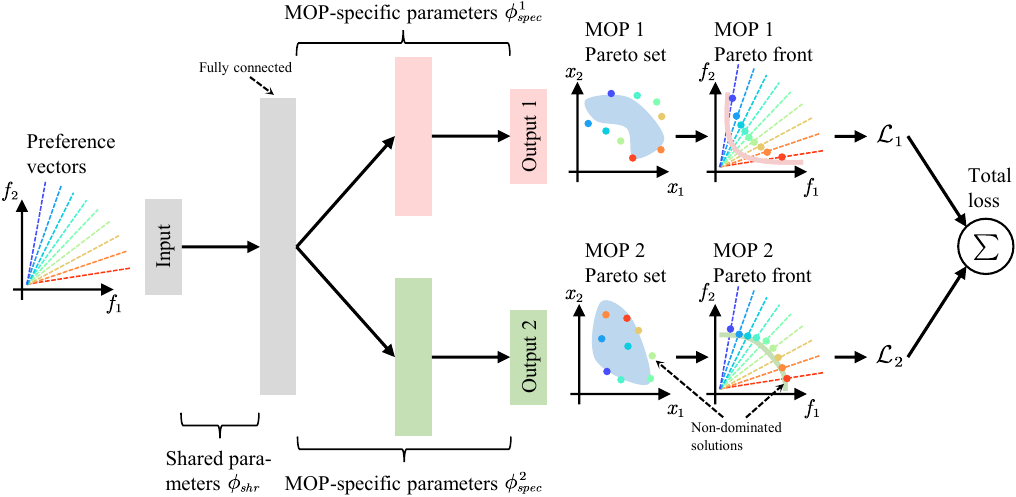}
    \caption{The fully connected neural network model of collaborative Pareto set learning, which is structured with shared and MOP-specific parameters. This allows it to concurrently learn from multiple MOPs with fewer parameters, enabling more efficient training and inference.}
    \label{fig:CoPSLModel}
\end{figure*}

We begin by outlining the basics of the CoPSL model (refer to Fig.~\ref{fig:CoPSLModel}). The architecture of the CoPSL model is inspired by the concept of hard parameter sharing in MTL. It aims to simultaneously learn the Pareto sets of multiple MOPs with fewer parameters, enabling more efficient training and inference. In the CoPSL model, the parameter set is divided into shared parameters and MOP-specific parameters, as illustrated in Fig.~\ref{fig:CoPSLModel}. Let $h_\phi(\cdot)$ denote the CoPSL model with parameters $\phi = \{ \phi_{shr}, \{\phi^i_{spec}\}_{i=1}^K \}$, where $\phi_{shr}$ represents shared parameters and $\phi^i_{spec}$ represents specific parameters for the $i$-th MOP ($i=1,\ldots,K$).

\begin{algorithm}[t]
    \SetCustomAlgoRuledWidth{0.45\textwidth}
    \SetAlgoLined
    \caption{Collaborative Pareto Set Learning}
    \label{alg:CoPSL}
    \KwIn{number of MOPs $K$, number of iterations $T$, batch size of sampled preferences $B$, learning rate $\eta$}
    Initialize parameters $\phi = \{ \phi_{shr}, \{\phi^i_{spec}\}_{i=1}^K \}$ \\
    Initialize a weight vector $\mathbf{w} = (w_1, \ldots, w_K)$ \\
    \For{$t=1,\ldots, T$} {
        \textcolor[rgb]{0.43,0.43,0.43}{\tcp{Generate solution sets of multiple MOPs}}
        Sample $B$ preference vectors $\{ \mathbf{p}^b \}_{b=1}^B$ with a Dirichlet distribution ${\rm Dir}(\bm{\alpha})$ \\
        Generate a set of solutions per MOP: $\Big\{ \{ \mathbf{x}_i(\mathbf{p}^b) \}_{b=1}^B\Big\}_{i=1}^K \gets h_\phi\Big(\{ \mathbf{p}^b \}_{b=1}^B\Big)$ \\
        \textcolor[rgb]{0.43,0.43,0.43}{\tcp{Compute the losses of the solution sets}}
        Compute loss per MOP: $\mathcal{L}_i \gets \frac{1}{B} \sum^B_{b=1} \ell_i( \mathbf{x}_i^b \mid \mathbf{p}^b)~\forall i\in[K]$ \\
        Compute total loss: $\mathcal{L}_{tot} \gets \sum_{i=1}^K w_i \mathcal{L}_i$ \\
        \textcolor[rgb]{0.43,0.43,0.43}{\tcp{Update the CoPSL model}}
        Update the parameters: $\phi \gets \phi - \eta\nabla_\phi \mathcal{L}_{tot}$ \\
    }
    \KwOut{CoPSL model $\big\{ \mathbf{x}_i(\mathbf{p}) \big\}^K_{i=1}=h_\phi(\mathbf{p})$}
\end{algorithm}

Algorithm~\ref{alg:CoPSL} presents the pseudo-code of the proposed CoPSL framework. To begin, the parameters $\phi$ is initialized along with a weight vector $\mathbf{w}=(w_1, \ldots, w_K)$ (Line 1 and 2). In our implementation, we set $w_i=1$ $\forall i$. Each iteration of CoPSL consists of the following three steps:
\vspace{5pt}

\noindent \textbf{Generate solution sets of multiple MOPs.} Firstly, a batch of $B$ preference vectors $\{ \mathbf{p}^b \}_{b=1}^B$ is sampled from a Dirichlet distribution with parameter vector $\bm{\alpha}$ (Line 5). The CoPSL model takes these preferences as input and projects them into a higher-dimensional space by shared parameters $\phi_{shr}$ to seek out the common relationships among MOPs. Subsequently, each set of MOP-specific parameters $\phi^i_{spec}$ maps these features to generate corresponding sets of solutions $\{ \mathbf{x}_i(\mathbf{p}^b) \}_{b=1}^B$ for their respective MOPs (Line 6).
\vspace{5pt}

\noindent \textbf{Compute the losses of the solution sets.} With the generated solution $\mathbf{x}_i(\mathbf{p}^b)$, we can calculate its corresponding objective vector $\mathbf{F_i}\big(\mathbf{x}_i(\mathbf{p}^b)\big)$. The objective vectors are crucial for evaluating the solutions. By leveraging these vectors, a loss function can be designed to accordingly update the model, thereby refining its ability to approximate the Pareto set more accurately. For that purpose, we explore four alternatives for CoPSL as follows:

\textit{CoPSL-LS:} CoPSL-LS employs the straightforward linear aggregation function~\cite{navon2021learning}, which is particularly effective for MOPs with convex Pareto fronts~\cite{boyd2004convex}. The loss formula is expressed as:
\begin{equation}
    \ell_i(\mathbf{x}_i \mid \mathbf{p})=\sum^m_{j=1} p_j f^i_j(\mathbf{x}_i)
\end{equation}

\textit{CoPSL-COSMOS:} An extension of CoPSL-LS, CoPSL-COSMOS incorporates a cosine similarity regularization to minimize the angle between the objective vector of the generated solution and the preference vector~\cite{ruchte2021scalable}. The loss function for CoPSL-COSMOS is:
\begin{equation}
    \ell_i(\mathbf{x}_i \mid \mathbf{p})=\sum^m_{j=1} p_j f^i_j(\mathbf{x}) + \gamma \cos(\mathbf{p}, \mathbf{F}_i(\mathbf{x}_i))
\end{equation}
where $\gamma>0$ is a penalty hyperparameter.

\textit{CoPSL-TCH:} This method utilizes the Tchebycheff aggregation function, which is designed to focus on the worst-case deviation from the ideal point, denoted as $\mathbf{z}^\ast_i$, for the $i$-th objective vector~\cite{tuan2024framework}. Its loss function is formulated as follows:
\begin{equation}
    \ell_i(\mathbf{x}_i \mid \mathbf{p}) = \max_{1\leq j\leq m}\{p_j(f^i_j(\mathbf{x}_i) - (z^\ast_{ij} - \epsilon)) \}
\end{equation}
where $\epsilon$ is a negligible positive value introduced to prevent the loss from becoming zero. Notably, $\mathbf{z}^\ast_i$ is dynamically updated based on the evaluated solutions during training.

\textit{CoPSL-MTCH:} CoPSL-MTCH adopts the modified Tchebycheff aggregation function as its loss function, as described in~\cite{lin2022pareto}. The formula is given by:
\begin{equation}
    \ell_i(\mathbf{x}_i \mid \mathbf{p}) = \max_{1\leq j\leq m}\{\frac{1}{p_j}(f^i_j(\mathbf{x}_i) - (z^\ast_{ij} - \epsilon)) \}
\end{equation}

Based on the aforementioned loss function $\ell_i(\cdot\mid\cdot)$, we can compute the loss $\mathcal{L}_i$ for the $i$-th MOP (Line 8). In our experiments detailed in Section~\ref{sec:exp_psl}, we conduct a comparative analysis of the four alternatives, both within and outside the CoPSL framework.

To evaluate the total loss across the solution sets of different MOPs, we constructed a total loss function $\mathcal{L}_{tot}$ in a manner akin to MTL (Line 9). It is defined as follows:
\begin{equation}
    \mathcal{L}_{tot} = \sum_{i=1}^K w_i \cdot \mathcal{L}_i
    = \sum_{i=1}^K w_i \cdot \frac{1}{B} \sum^B_{b=1} \ell_i(\mathbf{x}_i^b\mid \mathbf{p}^b)
\end{equation}
\vspace{5pt}

\noindent \textbf{Update the CoPSL model.} The CoPSL model parameters $\phi$ enable to be updated based on the total loss $\mathcal{L}_{tot}$ (Line 11). Particularly, the set of MOP-specific parameters $\phi^i_{spec}$ for the $i$-th MOP is updated according to the corresponding loss $\mathcal{L}_i$. This suggests that $\phi^i_{spec}$ is fine-tuned based on its associated specific solution set. The updated formula is given by:
\begin{equation}
    \phi^i_{spec} = \phi^i_{spec} - \eta \nabla_{\phi^i_{spec}} \mathcal{L}_i
\end{equation}
where $\eta$ denotes the learning rate. Regarding the shared parameters set $\phi_{shr}$, it is updated by the aggregated loss $\mathcal{L}_{tot} = \sum_{i=1}^K w_i\mathcal{L}_i$, indicating that $\phi_{shr}$ collaboratively learns from all the Pareto sets across various MOPs. The updated formula is as follows:
\begin{equation} \label{eq:shr}
    \phi_{shr} = \phi_{shr} - \eta \sum^K_{i=1} w_i \nabla_{\phi_{shr}} \mathcal{L}_i
\end{equation}

Notably, this formula elucidates the essence of the collaborative learning approach in CoPSL, specifically leveraging backpropagation informed by insights from multiple MOPs and thereby harnessing the possible interrelations.
\vspace{5pt}

Based on the above three steps, the CoPSL model achieves efficient training of multiple MOPs simultaneously in a collaborative fashion. \\

\noindent \textbf{CoPSL Scalability.} Equation~\eqref{eq:shr} reveals that when the gradients of MOPs are in conflict, or when the magnitude of one MOP gradient overshadows others, the resulting updates to the model may be suboptimal. In traditional MTL, various techniques have been proposed to balance the gradient magnitudes by dynamically adjusting the weight vector $\mathbf{w}$~\cite{chen2018gradnorm, liu2019end, guo2018dynamic}. Our approach allows us to easily incorporate these balancing techniques following the update of the CoPSL model (i.e., after Line 11 of Algorithm~\ref{alg:CoPSL}). Notably, MOPs enable dynamic adjustments using indicators in addition to loss functions (i.e., the aggregation functions)~\cite{wagner2007pareto,tian2017indicator}. These indicators, such as hypervolume (HV) and inverted generational distance (IGD), provide a more direct measure of the current state of the output solutions compared to loss functions alone. They offer supplementary information beyond losses to better guide the adjustment of the weight vector. We will discuss this as future work in Section~\ref{section6}. \\

\noindent \textbf{Differences with MTL.} The primary distinctions between MTL and CoPSL lie in the nature of their inputs and the setup of their shared layers. Specifically:
\begin{itemize}
    \item MTL typically processes image data as inputs. The image data for various tasks may be similar or distinctly different, but it is rich in information. In contrast, CoPSL utilizes preference vectors as inputs, which are essentially low-dimensional vectors of weights. These preference vectors for different MOPs are sampled in the same objective space, rendering them almost equivalent and less informative compared to image data.
    \item Hard parameter sharing models in MTL generally employ shared convolutional layers that are adept at handling image inputs. This setup facilitates the extraction of highly independent representations from the image data, which can be shared across tasks to enhance performance. On the other hand, CoPSL utilizes shared fully connected layers that receive the sampled preference vectors as inputs, as depicted in Fig.~\ref{fig:CoPSLModel}. It remains to be seen whether these fully connected layers are capable of learning the common relationships that can be beneficially shared among MOPs.
\end{itemize}

To provide a deeper understanding of CoPSL, we will delve into a more thorough analysis in the following section.

\section{Understanding Shared Representations in CoPSL} \label{section4}

\begin{figure*}[!t]
    \centering
    \includegraphics[scale=0.89]{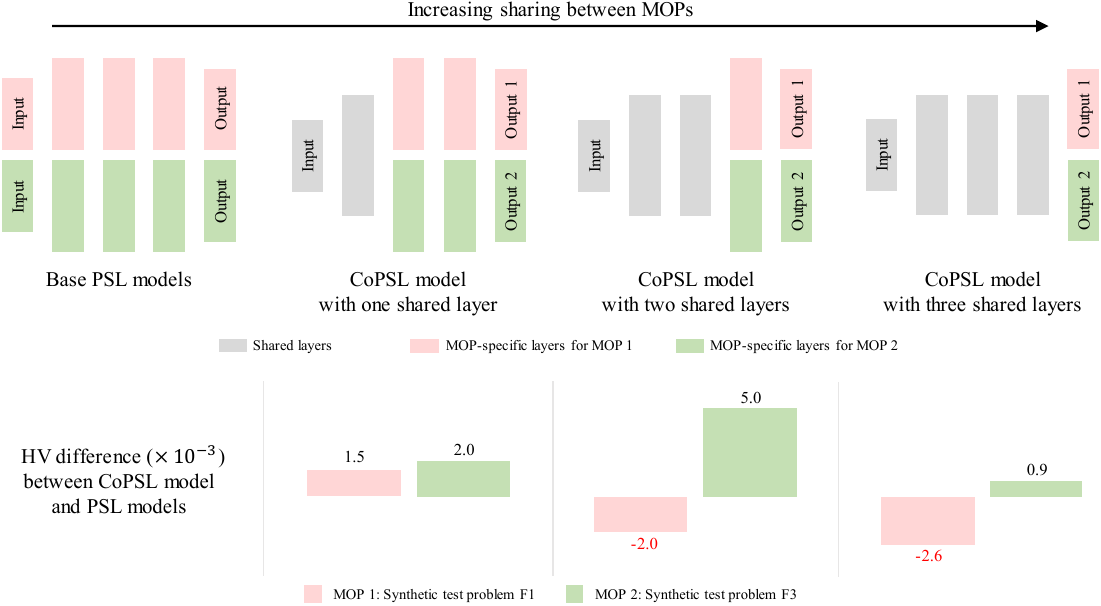}
    \caption{A toy example to explore mechanisms of the CoPSL model. We train several CoPSL model architectures, each sharing a different number of layers for two MOPs. For each of these models, we plot their performance on each MOP relative to the MOP-specific model. The results demonstrate that the presence of shareable representations for preference vectors is more effectively learned in a collaborative fashion.}
    \label{fig:UnderstandingCoPSL}
\end{figure*}

Given the differences between MTL and CoPSL, we aim to investigate whether CoPSL can learn shareable relationships across MOPs and deepen our understanding of CoPSL's mechanisms. To this end, we perform extensive experimental analysis. Consider a simple experiment where we train a variety of CoPSL model architectures with different numbers of shared layers, each designed to tackle two arbitrary MOPs (e.g., synthetic two-dimensional benchmarks F1 and F3~\cite{lin2022pareto}). The upper part of Fig.~\ref{fig:UnderstandingCoPSL} illustrates the spectrum of model architectures that were constructed for this purpose. At one extreme, we train two separate PSL models for each MOP, with no cross-talk between them. At the other extreme, we train one generic model where all layers are shared, and only the last layers are MOP-specific. We exhaustively enumerate all possible architectures as depicted in the upper part of Fig.~\ref{fig:UnderstandingCoPSL}, and evaluate their respective performance as shown in the lower part of Fig.~\ref{fig:UnderstandingCoPSL}.

The findings indicate that while there appears to be little correlation between the two MOPs F1 and F3, employing CoPSL models with one or two shared layers can still improve overall performance. It demonstrates that certain representations of preferences (inputs) can be generalized and shared across multiple MOPs, rather than being learned separately from scratch for each individual MOP. Notably, this occurs even when the Pareto sets (outputs) are irrelevant, revealing the existence of an effective shareable representation space for preferences across diverse MOPs with limited inherent relation. Leveraging shared knowledge across different MOPs is beneficial for better approximating the Pareto sets.

Furthermore, we note that introducing only one shared layer improves the HV metrics for both MOPs to a certain extent. However, increasing the number of shared layers beyond one reveals a `seesaw' pattern in the HV values for the respective MOPs. This can be attributed to two primary reasons. Firstly, the model's input consists merely of two-dimensional preference vectors. In cases where the dimensionality is low, the preference vectors present more straightforward shared representations that a single shared layer can effectively capture. Although additional shared layers have the potential to learn more complex representations, they tend to diminish the model's ability to map these representations to Pareto solutions. Secondly, the disparity in Pareto sets among various MOPs generates conflicting gradients. As delineated in Equation~\eqref{eq:shr}, this conflict adversely affects the updates of shared parameters $\phi_{shr}$. Consequently, as previously mentioned, the given weight vectors face inherent limitations in optimizing for both MOPs simultaneously.

Given that only two MOPs are considered in this section, we will further explore the effectiveness of CoPSL by simultaneously tackling a greater number of MOPs in Section~\ref{sec:exp_psl}, including a suite of five to six MOPs.

\section{Experiments} \label{section5}
\begin{table*}[t]
\centering
\caption{Comparisons of EMO algorithms, PSL approaches, and our proposed algorithm on a suite of six synthetic benchmarks. The best results are highlighted in bold font. CoPSL not only runs significantly faster than the compared algorithms, but also slightly outperforms them in terms of approximation capability.}
\label{tab:exp_emo}
\begin{tabular}{lccccccccc}
\toprule 
         & \multicolumn{6}{c}{HV $\uparrow$}                                          & \multirow{2}{*}{\makecell[c]{Runtime\\(s) $\downarrow$}} & \multirow{2}{*}{\makecell[c]{FLOPs\\(K) $\downarrow$}} & \multirow{2}{*}{\makecell[c]{Params\\(K) $\downarrow$}} \\
\cmidrule{2-7}
         & F1       & F2       & F3       & F4       & F5       & F6       &                           &                        &                         \\

\midrule
NSGA-II  & 7.957e-01 & 7.922e-01 & 7.835e-01 & 7.843e-01 & 7.832e-01 & 8.098e-01 & 9.771e+00                  & -                       & -                        \\
NSGA-III  & 8.089e-01 & 8.106e-01 & \textbf{8.165e-01} & 8.014e-01 & 8.139e-01 & 8.295e-01 & 9.502e+00        & -                       & -                        \\
MOEA/D   & 8.175e-01 & 8.192e-01 & 8.164e-01 & 8.023e-01 & 8.060e-01 & 8.276e-01 & 2.868e+01                  & -                       & -                        \\
PSL-TCH  & 8.578e-01 & 8.651e-01 & 8.002e-01 & 8.235e-01 & 8.299e-01 & 8.456e-01 & 1.003e+01                  & 8.110e+02                       & 4.086e+02                        \\
CoPSL-TCH & \textbf{8.600e-01} & \textbf{8.651e-01} & 8.019e-01 & \textbf{8.247e-01} & \textbf{8.306e-01} & \textbf{8.465e-01} & \textbf{8.364e+00} & \textbf{8.059e+02}  & \textbf{4.048e+02}         \\
\bottomrule
\end{tabular}
\end{table*}

\begin{figure*}[!t]
    \centering
    \subfigure[F1]{
    \begin{minipage}[b]{0.3\linewidth}
    \centering
    \includegraphics[scale=0.255]{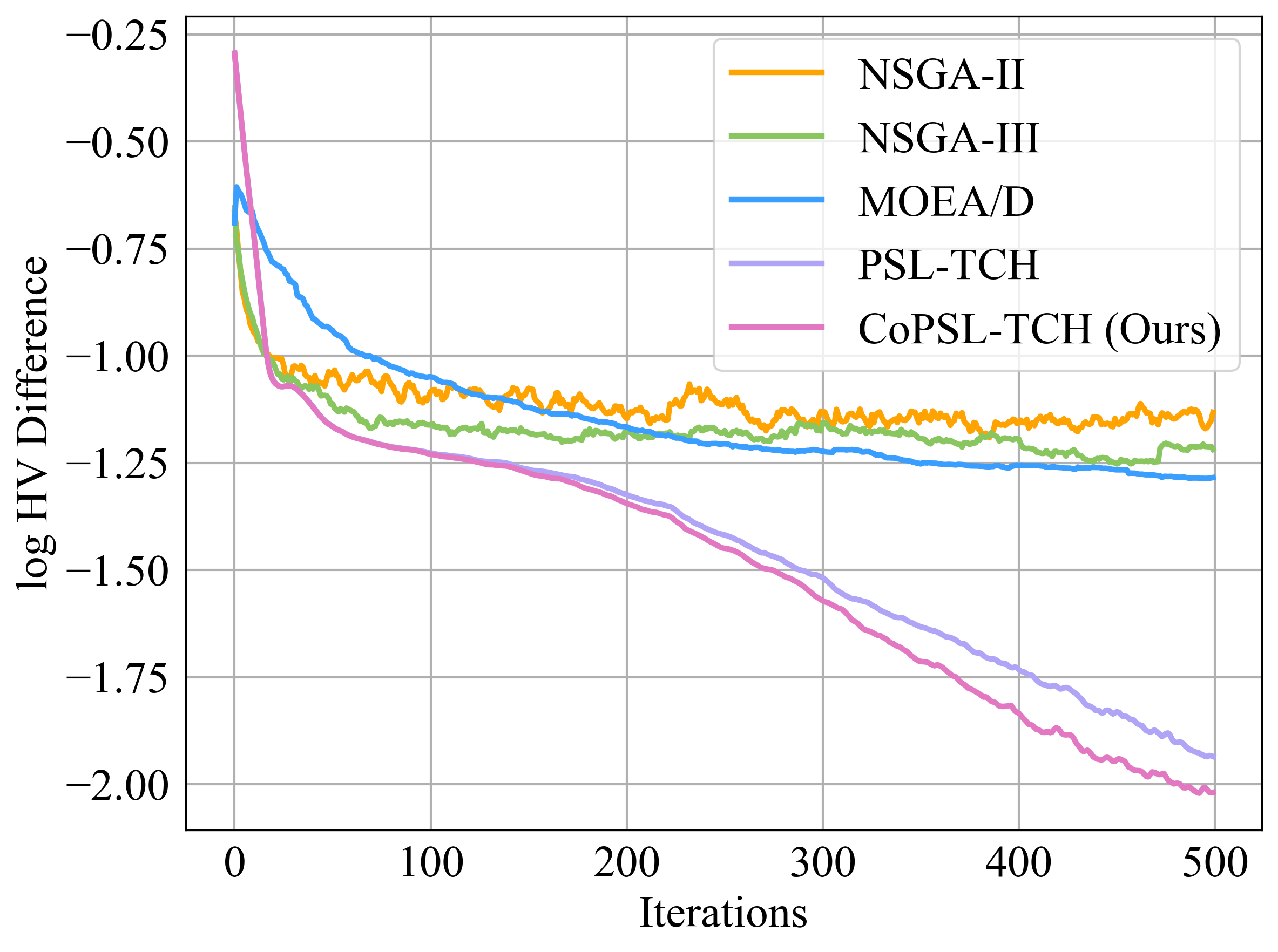}
    \end{minipage}
    }
    \subfigure[F2]{
    \begin{minipage}[b]{0.3\linewidth}
    \centering
    \includegraphics[scale=0.255]{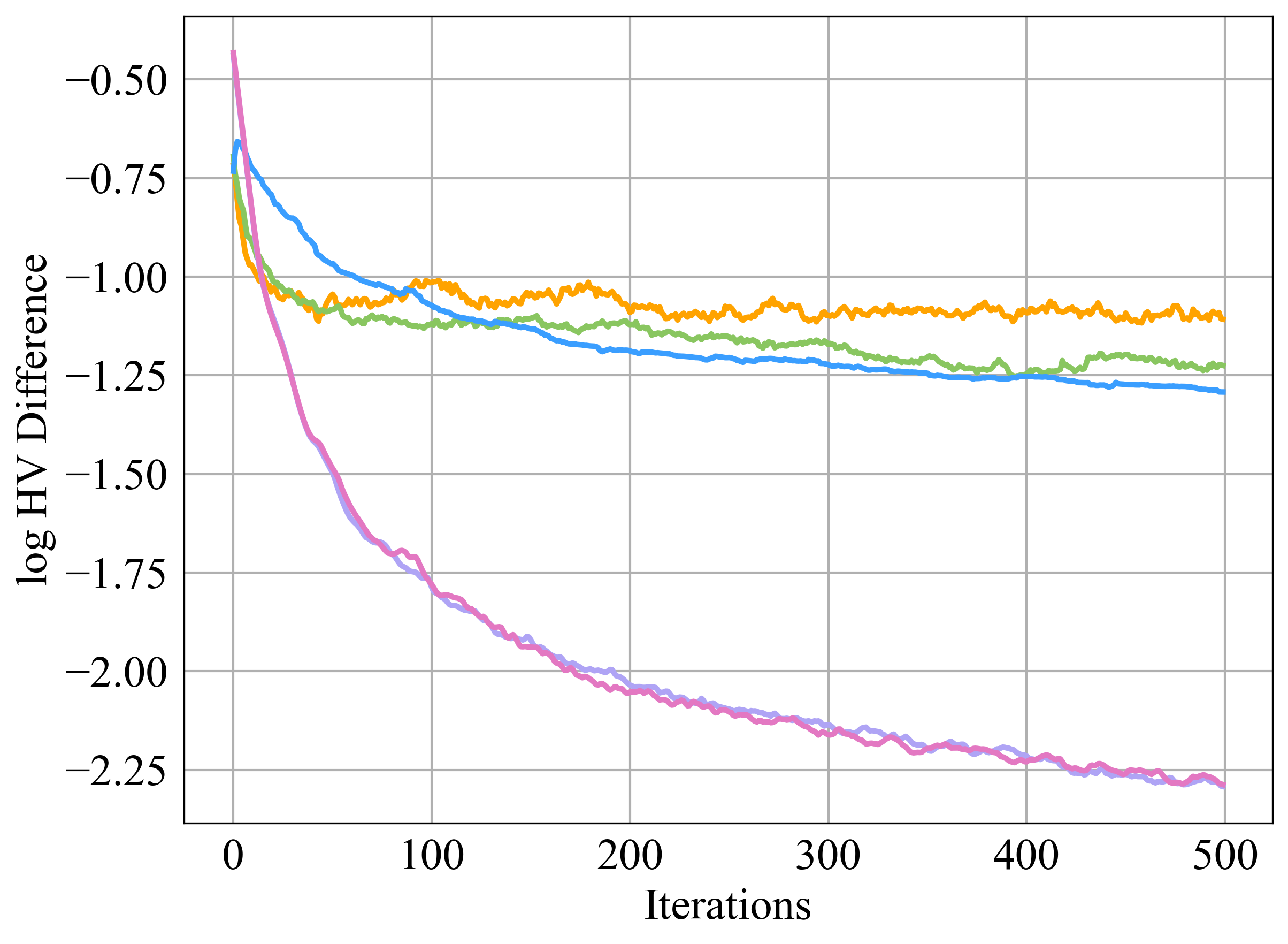}
    \end{minipage}
    }
    \subfigure[F3]{
    \begin{minipage}[b]{0.3\linewidth}
    \centering
    \includegraphics[scale=0.255]{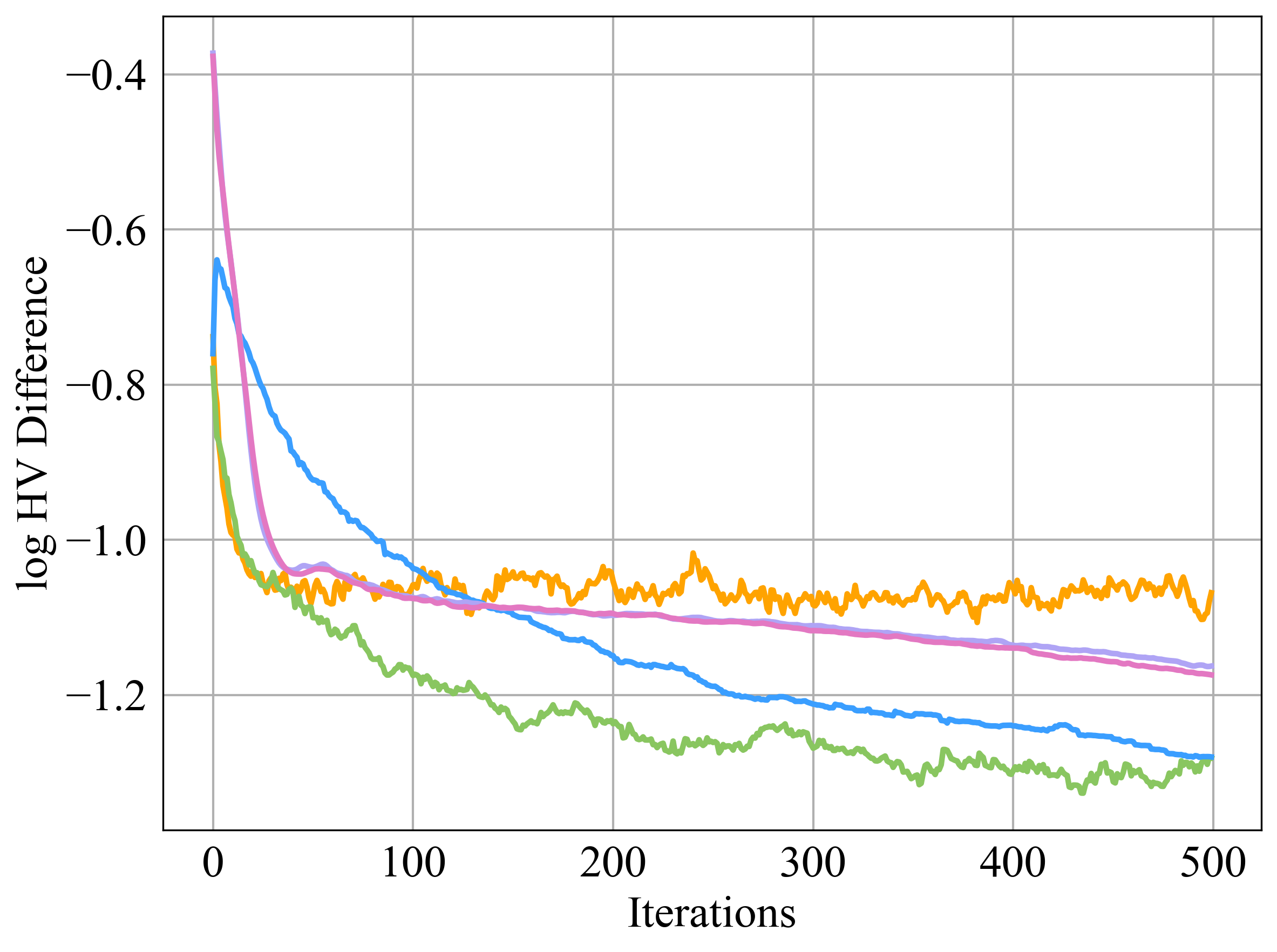}
    \end{minipage}
    }
    \subfigure[F4]{
    \begin{minipage}[b]{0.3\linewidth}
    \centering
    \includegraphics[scale=0.255]{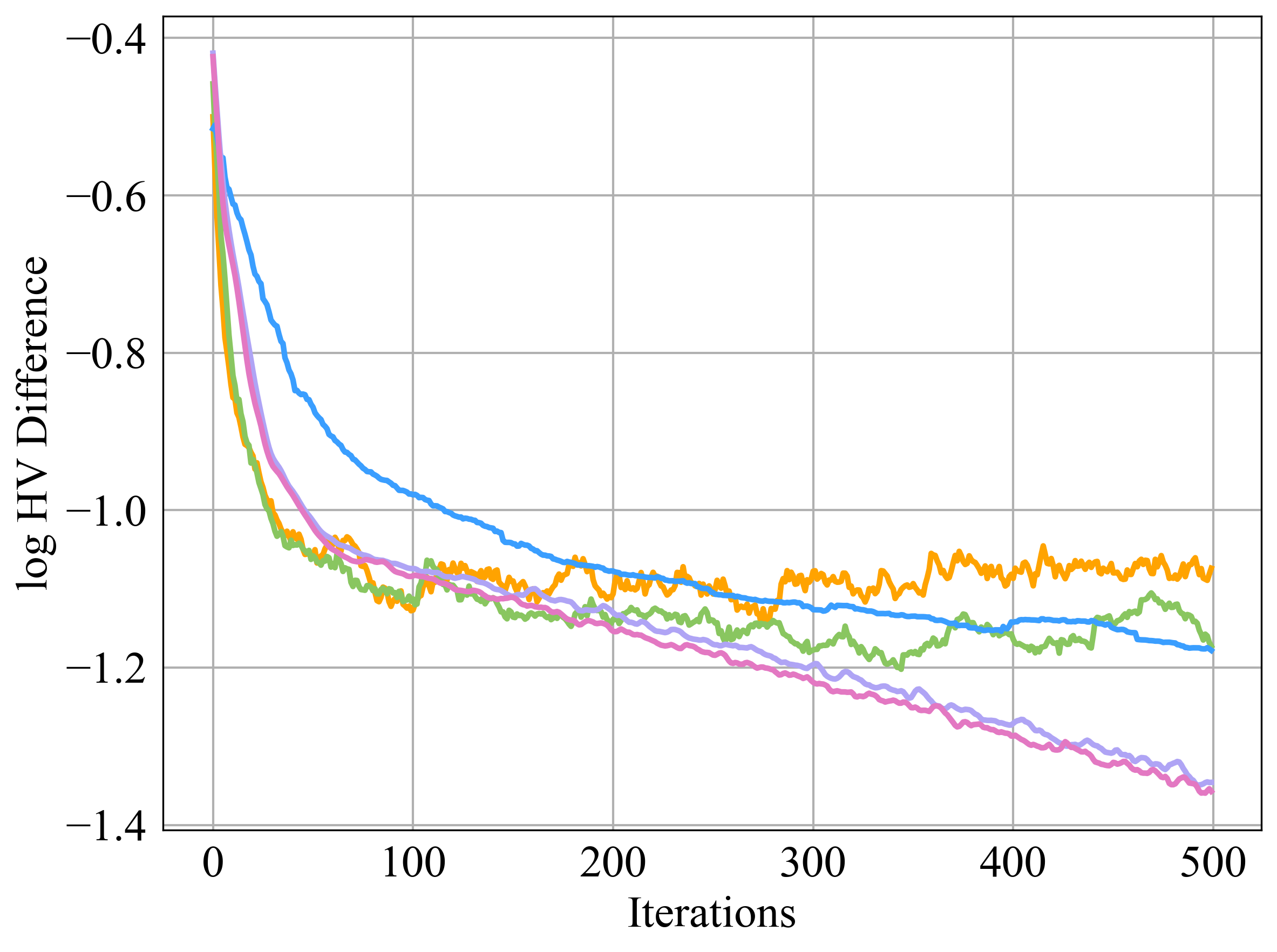}
    \end{minipage}
    }
    \subfigure[F5]{
    \begin{minipage}[b]{0.3\linewidth}
    \centering
    \includegraphics[scale=0.255]{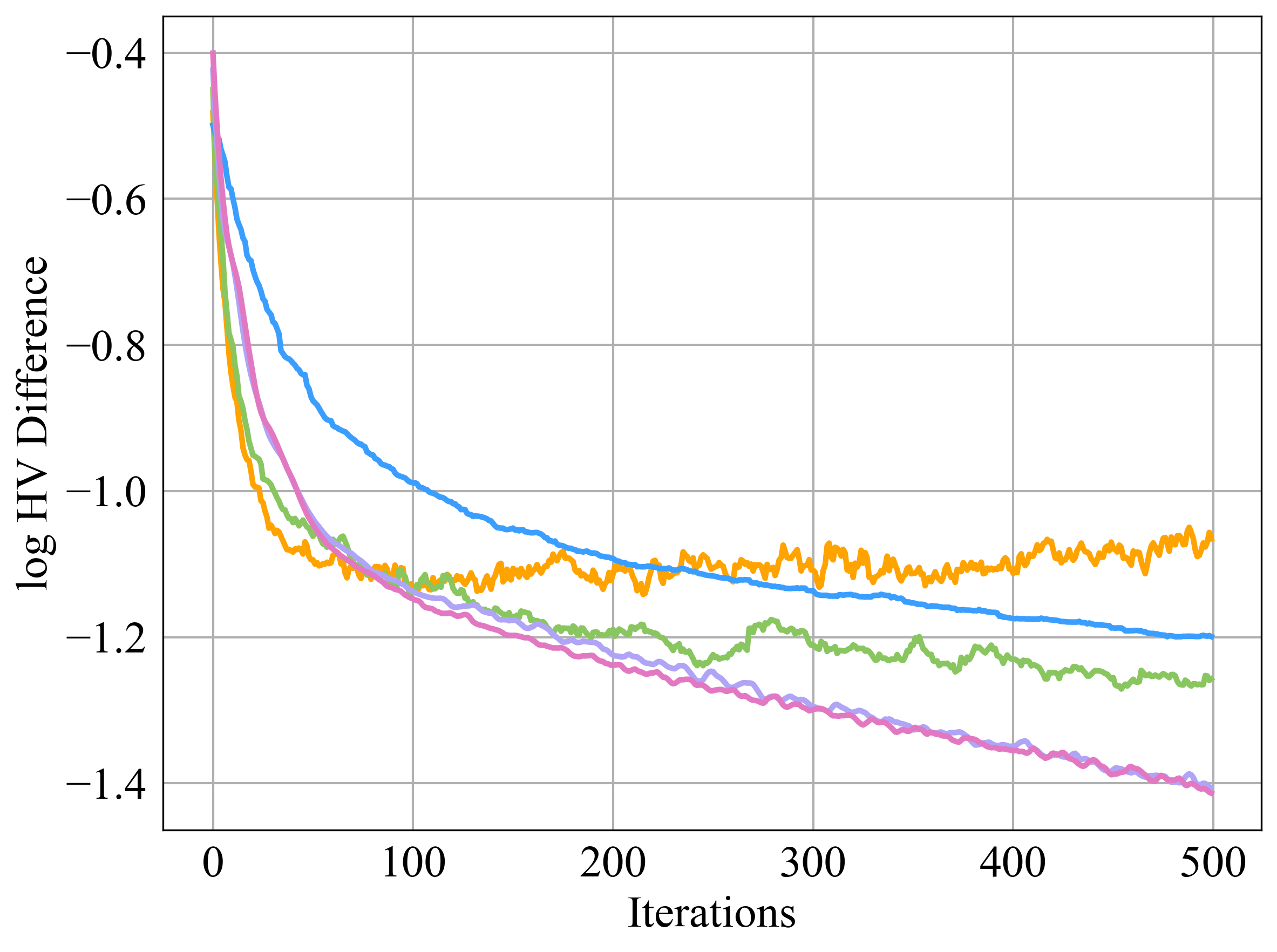}
    \end{minipage}
    }
    \subfigure[F6]{
    \begin{minipage}[b]{0.3\linewidth}
    \centering
    \includegraphics[scale=0.255]{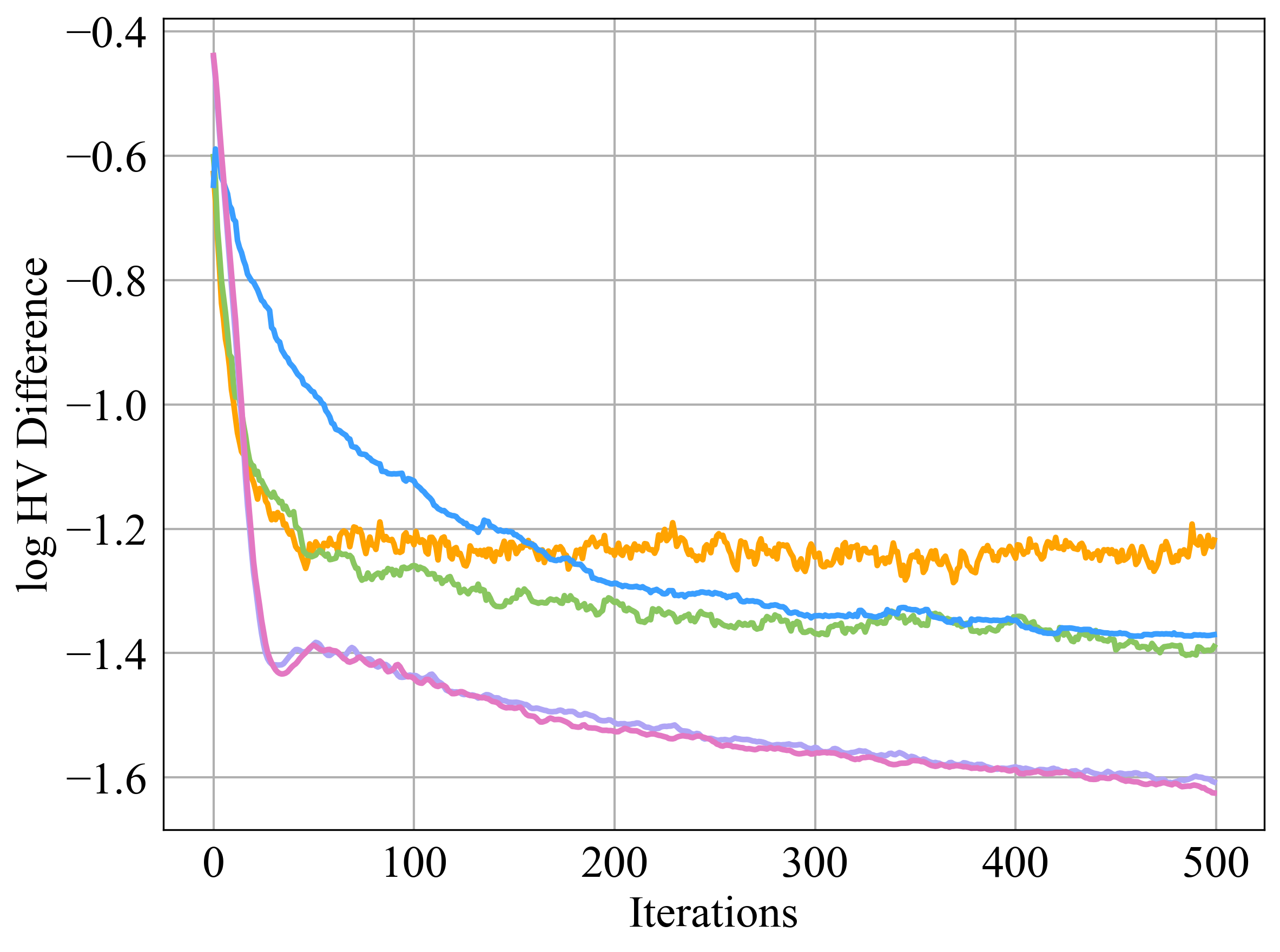}
    \end{minipage}
    }
    \caption{The log HV difference curves of four algorithms for six different problems. The solid line is the mean value averaged over 10 independent runs for each algorithm. The labels of the algorithms can be found in Subfig. (a).} \label{fig:exp_emo}
\end{figure*}

\subsection{Experimental Setups}

\noindent \textbf{Baselines.} We consider three popular model-free methods and four state-of-the-art model-based approaches as baselines. \textit{1) EMO algorithms:} NSGA-II~\cite{deb2002fast}, NSGA-III~\cite{deb2013evolutionary}, MOEA/D~\cite{zhang2007moea}, which are implemented using pymoo. \textit{2) PSL algorithms:} PSL-LS~\cite{navon2021learning}, COSMOS~\cite{ruchte2021scalable}, PSL-TCH~\cite{tuan2024framework}, PSL-MTCH~\cite{lin2022pareto}, which are executed in Pytorch. 
\vspace{5pt}

\noindent \textbf{Benchmarks and Real-World Problems.} The algorithms are first compared on a suite of six synthetic two-dimensional test problems, including F1 to F6 in~\cite{lin2022pareto}. Then experiments are also conducted on five different three-dimensional real-world multi-objective engineering design problems~\cite{tanabe2020easy}, including 1) Two bar truss design (RE31); 2) Welded beam design (RE32); 3) Disc brake design (RE33); 4) Vehicle crashworthiness design (RE34); 5) Rocket injector design (RE37). Comprehensive details regarding these problems are provided in the Appendix.

Notably, in contrast to non-collaborative approaches that run separate populations/models for each problem, CoPSL concurrently processes all problems from a given problem suite in the same experimental run using a single model. Additionally, due to limitations within the inputs of the CoPSL model, the number of objective dimensions remained consistent across all problems in each set of experiments.
\vspace{5pt}

\noindent \textbf{Evaluation Metrics.} The evaluation of algorithmic performance focuses on three primary aspects. \textit{1) Approximation quality of the Pareto set.} We employ the HV indicator, which comprehensively evaluates the diversity and convergence of the solution set. For a nuanced analysis, we further examine the log HV difference (i.e., the logarithm of the difference between the HV value of the true/approximate PF and the learned PF) to evaluate the gap between the true/approximate and the learned Pareto front. \textit{2) Algorithmic efficiency.} We measure this in terms of the actual runtime required to process a problem suite, which serves as a practical performance metric. For model-based algorithms, we further delve into the computational time complexity by analyzing Floating-point Operations (FLOPs), which offers a theoretical perspective on the computational effort required. \textit{3) Space complexity of algorithms.} Regarding model-based algorithms, we consider the total number of parameters within the model, denoted as Params, as a measure of the theoretical computational space complexity.
\vspace{5pt}

\noindent \textbf{Parameter settings.} For each experiment, all algorithms are executed 10 times independently. Specifically, we set the number of iterations $T$ to 500 and utilize the Adam optimizer~\cite{kingma2014adam} with its default parameters for model-based algorithms. The learning rate $\eta$ is set to $10^{-3}$ and the hyperparameter $\gamma$ in COSMOS is set to 100. Additionally, for our proposed approach, we fix the weight vector $\mathbf{w}$ as an all-one vector. The detailed exposition of the model architecture is provided in the Appendix.

\begin{table*}[t]
\centering
\caption{Comparisons of PSL approaches and CoPSL on six synthetic problems. The best results are highlighted in bold font. CoPSL not only runs significantly faster than its counterparts, but also slightly outperforms them in approximation capability.}
\label{tab:exp_PSL_F}
\begin{tabular}{lccccccc}
\toprule
            & \multicolumn{6}{c}{HV $\uparrow$}                                                                                     & \multirow{2}{*}{Runtime (s) $\downarrow$} \\
            \cmidrule{2-7}
            & F1                & F2                & F3                & F4                & F5                & F6                &                          \\
\midrule
PSL-LS      & 8.025e-01          & 8.579e-01          & \textbf{7.563e-01} & 7.934e-01          & 8.045e-01          & 8.347e-01          & 9.823e+00                 \\
CoPSL-LS     & \textbf{8.305e-01} & \textbf{8.586e-01} & 7.522e-01          & \textbf{7.950e-01} & \textbf{8.072e-01} & \textbf{8.355e-01} & \textbf{8.346e+00}        \\
\midrule
PSL-COSMOS  & 3.758e-01          & 1.921e-01          & 2.004e-01          & 2.622e-01          & 4.362e-01          & \textbf{1.302e-01} & 1.078e+01                 \\
CoPSL-COSMOS & \textbf{3.971e-01} & \textbf{2.240e-01} & \textbf{2.080e0-01} & \textbf{3.149e-01} & \textbf{4.384e-01} & 1.256e-01          & \textbf{9.207e+00}        \\
\midrule
PSL-TCH     & 8.595e-01          & \textbf{8.660e-01} & 8.035e-01          & 8.266e-01          & 8.330e-01          & 8.467e-01          & 1.010e+01                 \\
CoPSL-TCH    & \textbf{8.605e-01} & 8.654e-01          & \textbf{8.051e-01} & \textbf{8.287e-01} & \textbf{8.341e-01} & \textbf{8.472e-01} & \textbf{8.491e+00}        \\
\midrule
PSL-MTCH    & 8.249e-01          & 8.676e-01          & 7.881e-01          & 7.836e-01          & 7.947e-01          & 8.444e-01          & 1.028e+01                 \\
CoPSL-MTCH   & \textbf{8.326e-01} & \textbf{8.681e-01} & \textbf{7.910e-01} & \textbf{7.850e-01} & \textbf{8.019e-01} & \textbf{8.456e-01} & \textbf{8.638e+00}        \\
\bottomrule
\end{tabular}
\end{table*}

\begin{table*}[t]
\centering
\caption{Comparisons of PSL approaches and CoPSL on five real-world problems. The best results are highlighted in bold font. CoPSL not only runs significantly faster than its counterparts, but also marginally outperforms them in approximation capability.}
\label{tab:exp_PSL_RE}
\begin{tabular}{lccccccc}
\toprule
            & \multicolumn{5}{c}{HV $\uparrow$}                                                                                                               & \multirow{2}{*}{Runtime (s) $\downarrow$}    \\
            \cmidrule{2-6}
            & \multicolumn{1}{c}{RE31} & \multicolumn{1}{c}{RE32} & \multicolumn{1}{c}{RE33} & \multicolumn{1}{c}{RE34} & \multicolumn{1}{c}{RE37} & \multicolumn{1}{c}{}                         \\
\midrule
PSL-LS      & 1.325e+00                 & 1.200e+00                 & 5.821e-01                 & 4.085e-02                 & \textbf{9.087e-01}        & 8.448e+00                                     \\
CoPSL-LS     & \textbf{1.326e+00}        & \textbf{1.223e+00}        & \textbf{5.862e-01}        & \textbf{4.085e-02}        & 8.991e-01                 & \textbf{6.747e+00}                            \\
\midrule
PSL-COSMOS  & 1.327e+00                 & 1.150e+00                 & 5.183e-01                 & 4.085e-02                 & 7.130e-01                 & 8.871e+00                                     \\
CoPSL-COSMOS & \textbf{1.328e+00}        & \textbf{1.208e+00}        & \textbf{5.580e-01}        & \textbf{4.085e-02}        & \textbf{8.136e-01}        & \textbf{7.535e+00}                            \\
\midrule
PSL-TCH     & 1.325e+00                 & 1.211e+00                 & \textbf{6.925e-01}        & 4.085e-02                 & \textbf{1.030e+00}        & 8.496e+00                                     \\
CoPSL-TCH    & \textbf{1.326e+00}        & \textbf{1.235e+00}        & 6.871e-01                 & \textbf{4.085e-02}        & 1.022e+00                 & \textbf{6.949e+00}                            \\
\midrule
PSL-MTCH    & 1.325e+00                 & 1.134e+00                 & 6.358e-01                 & 4.085e-02                 & 1.025e+00                 & 8.566e+00                                     \\
CoPSL-MTCH   & \textbf{1.330e+00}        & \textbf{1.222e+00}        & \textbf{6.800e-01}        & \textbf{4.085e-02}        & \textbf{1.027e+00}        & \textbf{7.052e+00}                            \\
\bottomrule
\end{tabular}
\end{table*}

\begin{figure*}[!t]
    \centering
    \subfigure[F1]{
    \begin{minipage}[b]{0.3\linewidth}
    \centering
    \includegraphics[scale=0.255]{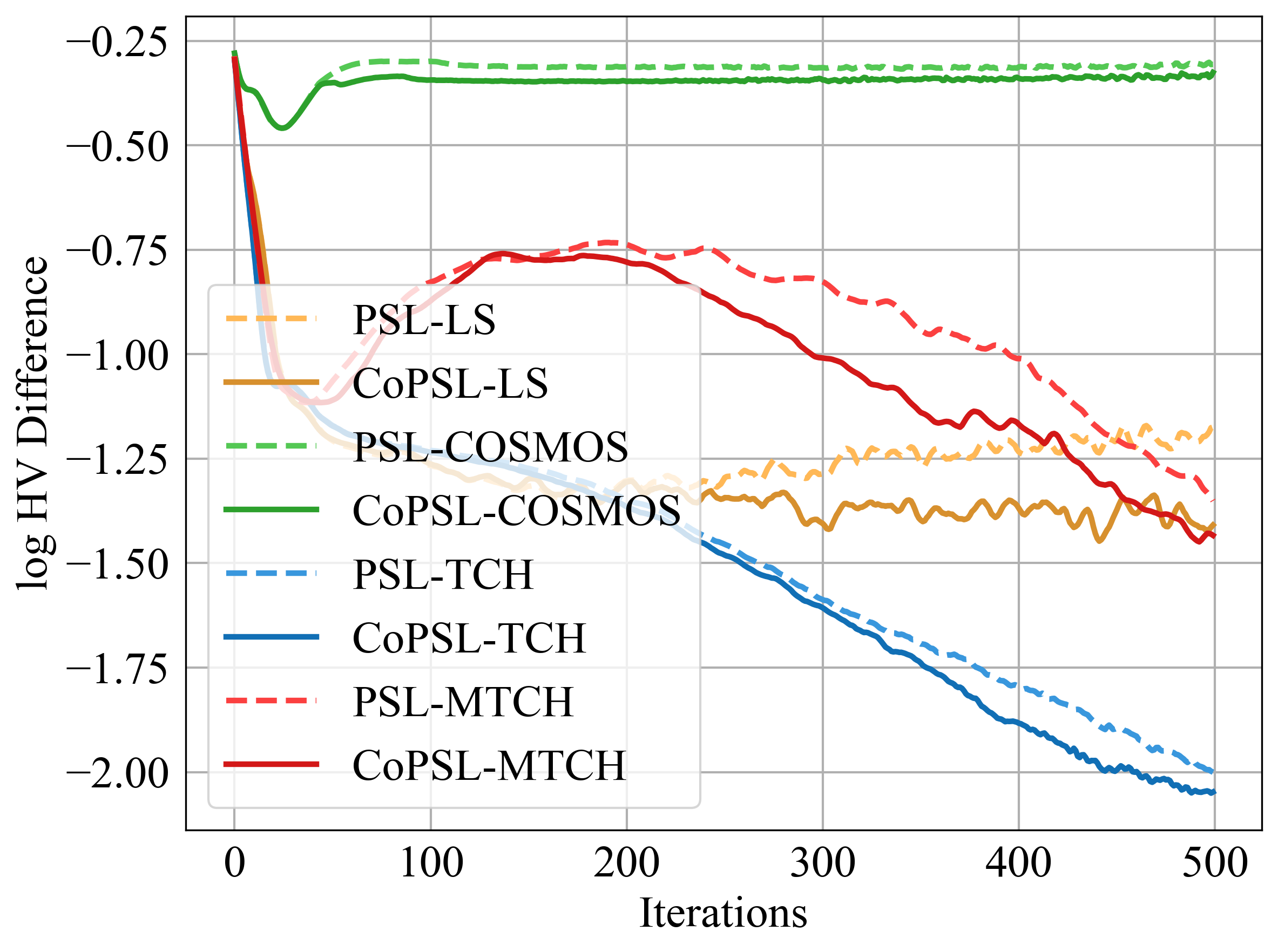}
    \end{minipage}
    }
    \subfigure[Two Bar Truss Design (RE31)]{
    \begin{minipage}[b]{0.3\linewidth}
    \centering
    \includegraphics[scale=0.255]{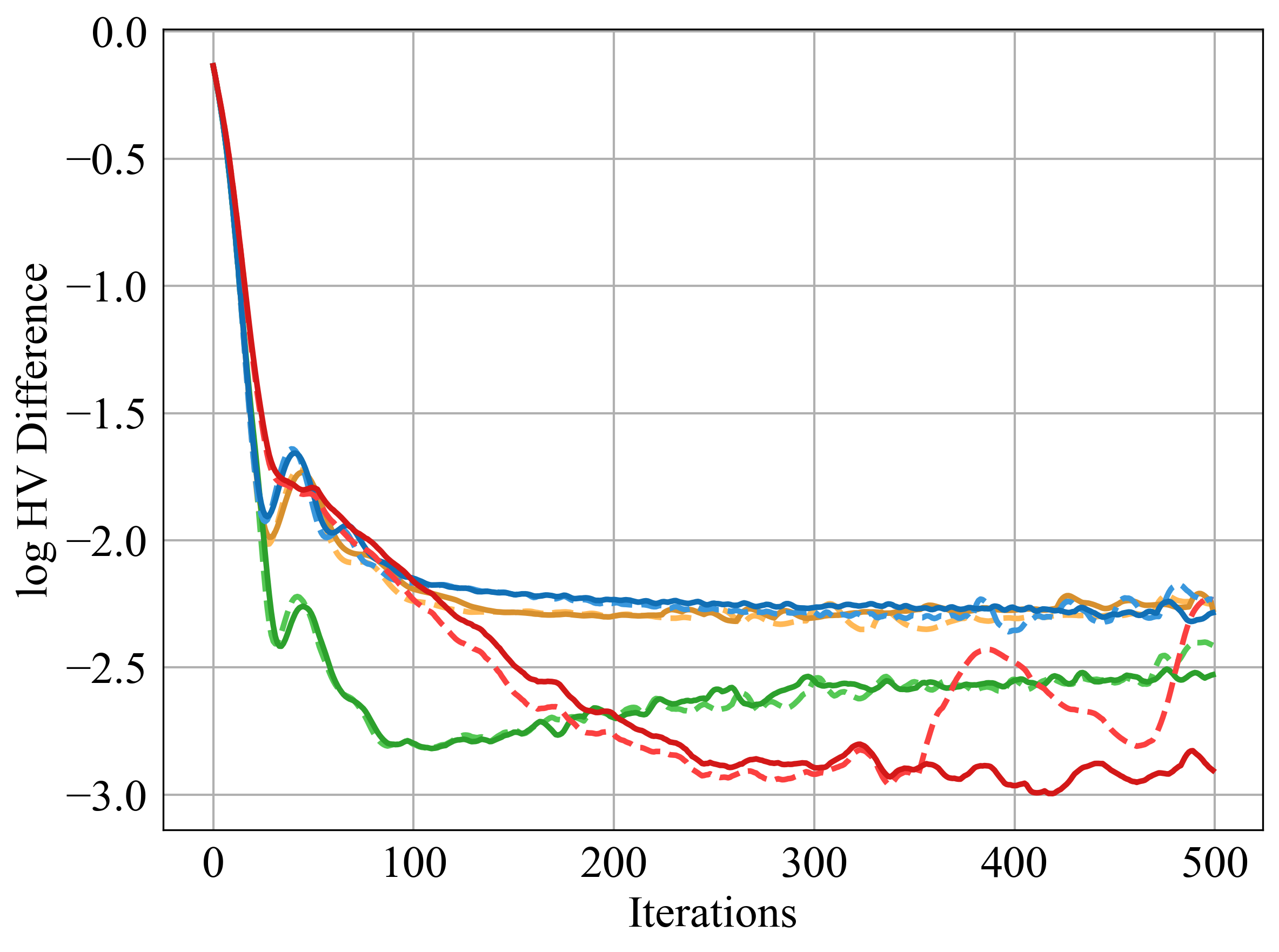} \label{fig:exp_psl_re32}
    \end{minipage}
    }
    \subfigure[Welded Beam Design (RE32)]{
    \begin{minipage}[b]{0.3\linewidth}
    \centering
    \includegraphics[scale=0.255]{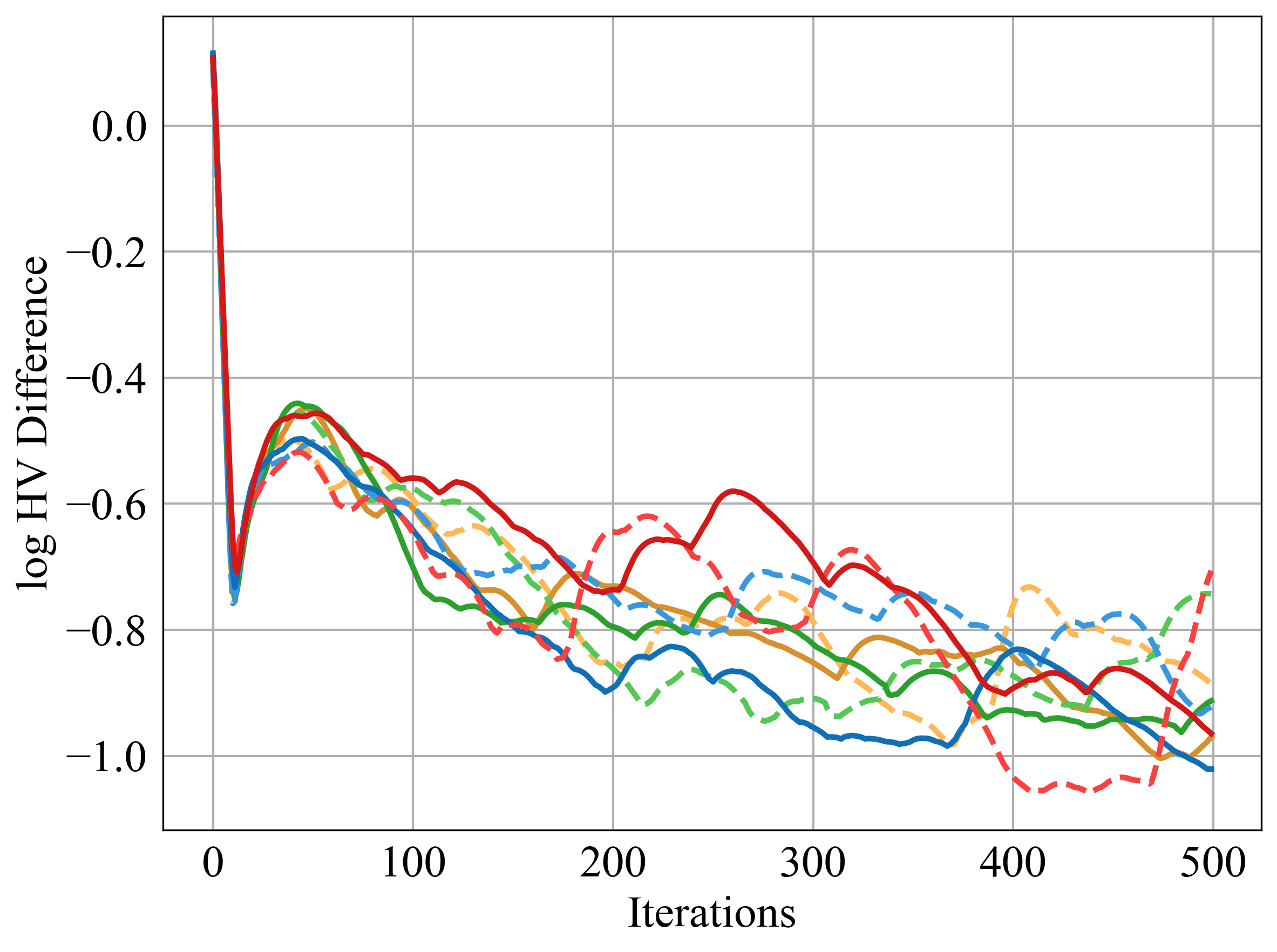}
    \end{minipage}
    }
    \caption{The log HV difference curves of four algorithms for three problems over 10 independent runs. The labels of the algorithms can be found in Subfig.~(a).} \label{fig:exp_psl}
\end{figure*}

\subsection{Comparisons with Base Methods}

To demonstrate the efficacy of our proposed methodology, we first conducted comparative analyses against traditional EMO algorithms and an advanced PSL algorithm on a suite of synthetic benchmarks (i.e., F1 to F6). For these experiments, we standardized the population size for EMOs and the batch size $B$ for model-based algorithms to 15. The results, detailed in Table~\ref{tab:exp_emo}, span multiple performance metrics of these methods, while Fig.~\ref{fig:exp_emo} illustrates the log HV difference to the approximate Pareto front during the optimization process.

CoPSL exhibits superior efficiency compared to all the baselines and demonstrates slightly better performance overall in terms of approximation capability. To be specific, CoPSL significantly outperforms the comparison algorithms in runtime performance and boasts lower theoretical FLOPs and Params than its model-based counterparts. This underscores the efficiency of CoPSL in concurrently learning multiple MOPs. Furthermore, CoPSL shows improvement over all the baselines in all but one MOP (i.e., F3), where it still maintains a slight edge over PSL. As depicted in Fig.~\ref{fig:exp_emo}, CoPSL converges slightly faster than the traditional single-MOP PSL approach. This indicates that CoPSL can effectively enhance both the approximation of the Pareto set and the robustness of the model. Additionally, it can be observed that EMOs can outpace traditional PSL in runtime when running with small population sizes. CoPSL addresses this limitation, extending the upper bounds of model-based Pareto set learning.

\subsection{Comparisons with SOTA PSL Methods} \label{sec:exp_psl}

\begin{figure*}[!t]
    \centering
    \subfigure[PSL-LS]{
    \begin{minipage}[b]{0.23\linewidth}
    \centering
    \includegraphics[scale=0.21]{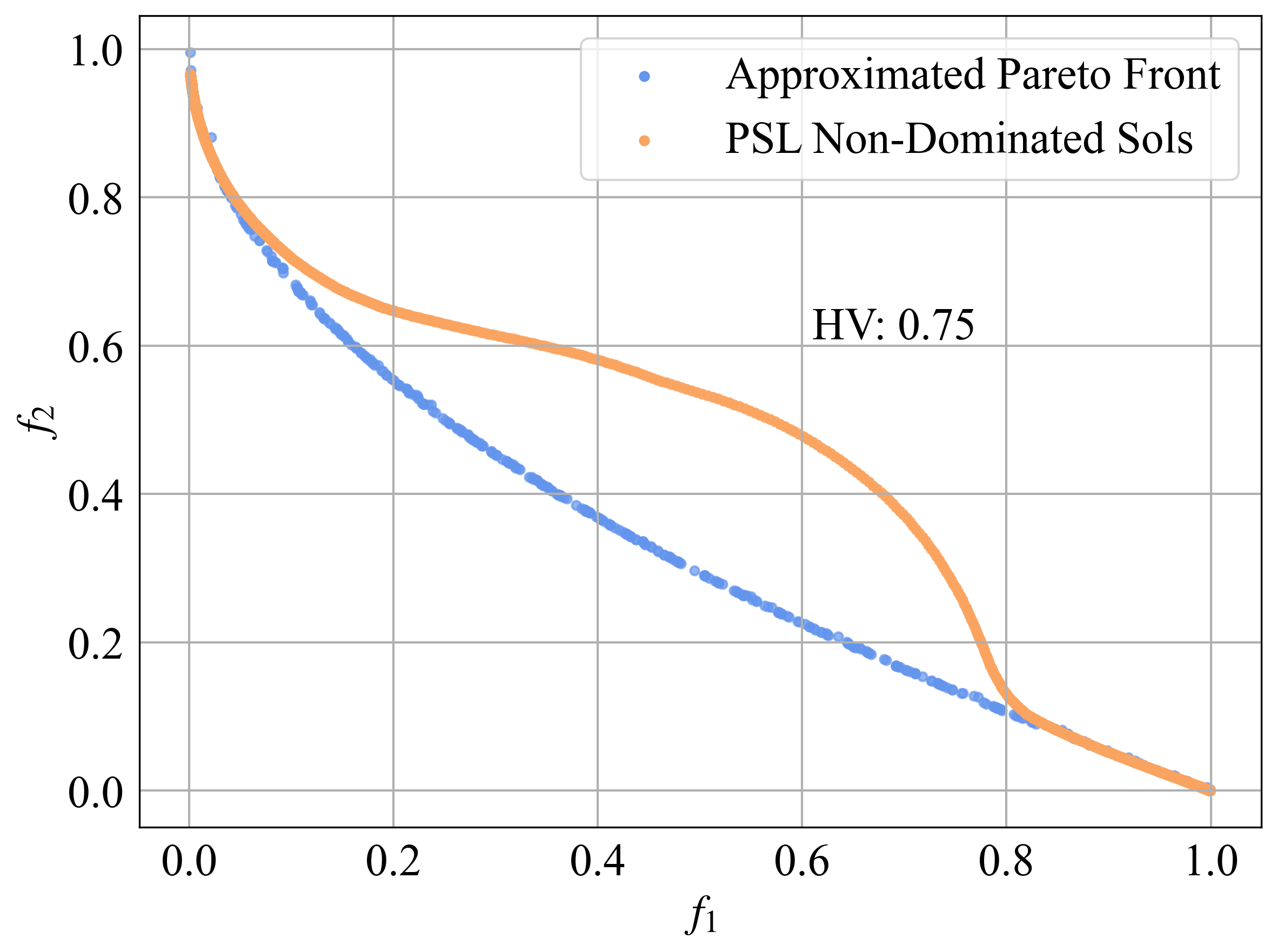}
    \end{minipage}
    }
    \subfigure[PSL-COSMOS]{
    \label{pslcosmos2}
    \begin{minipage}[b]{0.23\linewidth}
    \centering
    \includegraphics[scale=0.21]{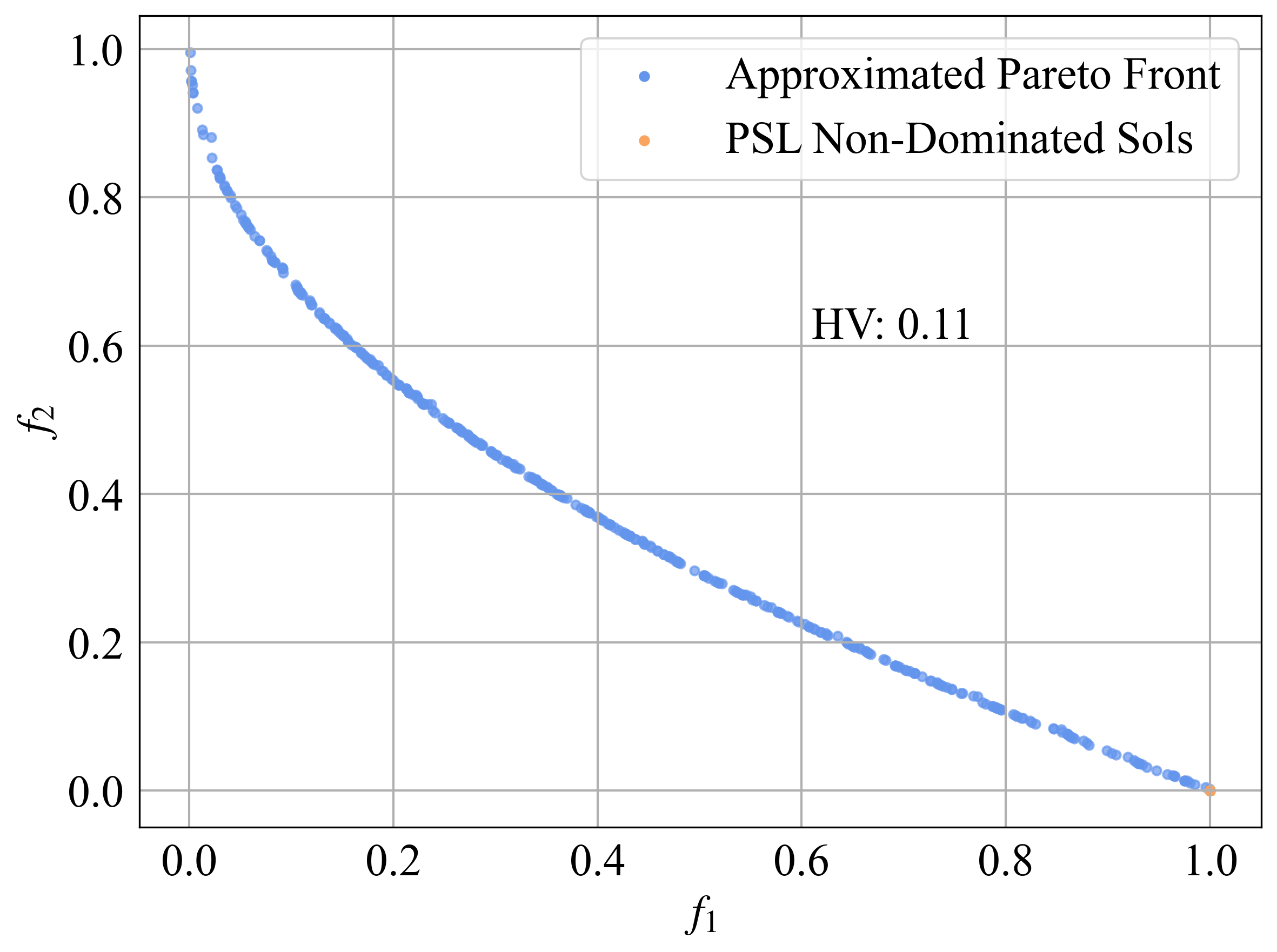}
    \end{minipage}
    }
    \subfigure[PSL-TCH]{
    \begin{minipage}[b]{0.23\linewidth}
    \centering
    \includegraphics[scale=0.21]{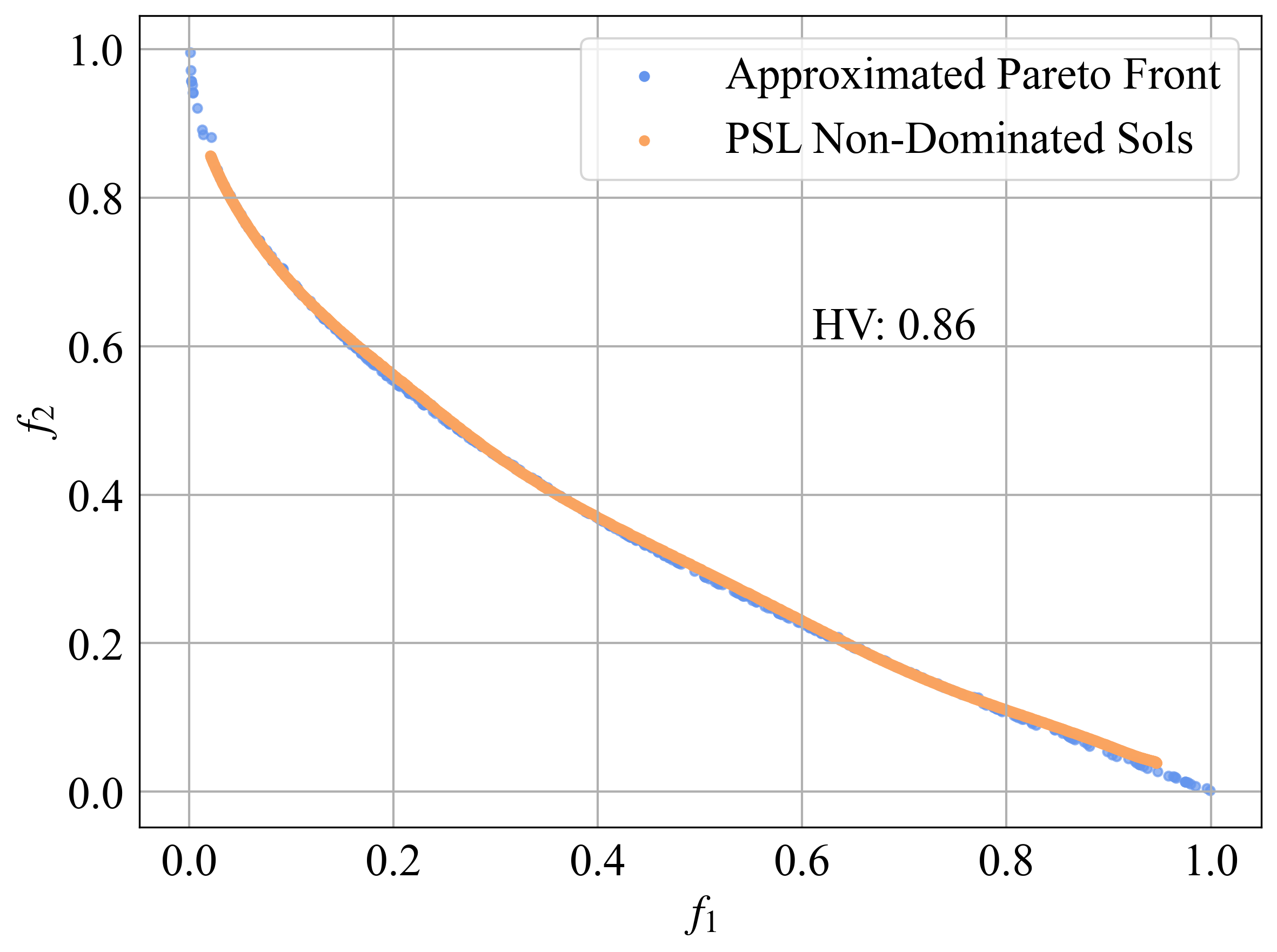}
    \end{minipage}
    }
    \subfigure[PSL-MTCH]{
    \label{pslmtch2}
    \begin{minipage}[b]{0.23\linewidth}
    \centering
    \includegraphics[scale=0.21]{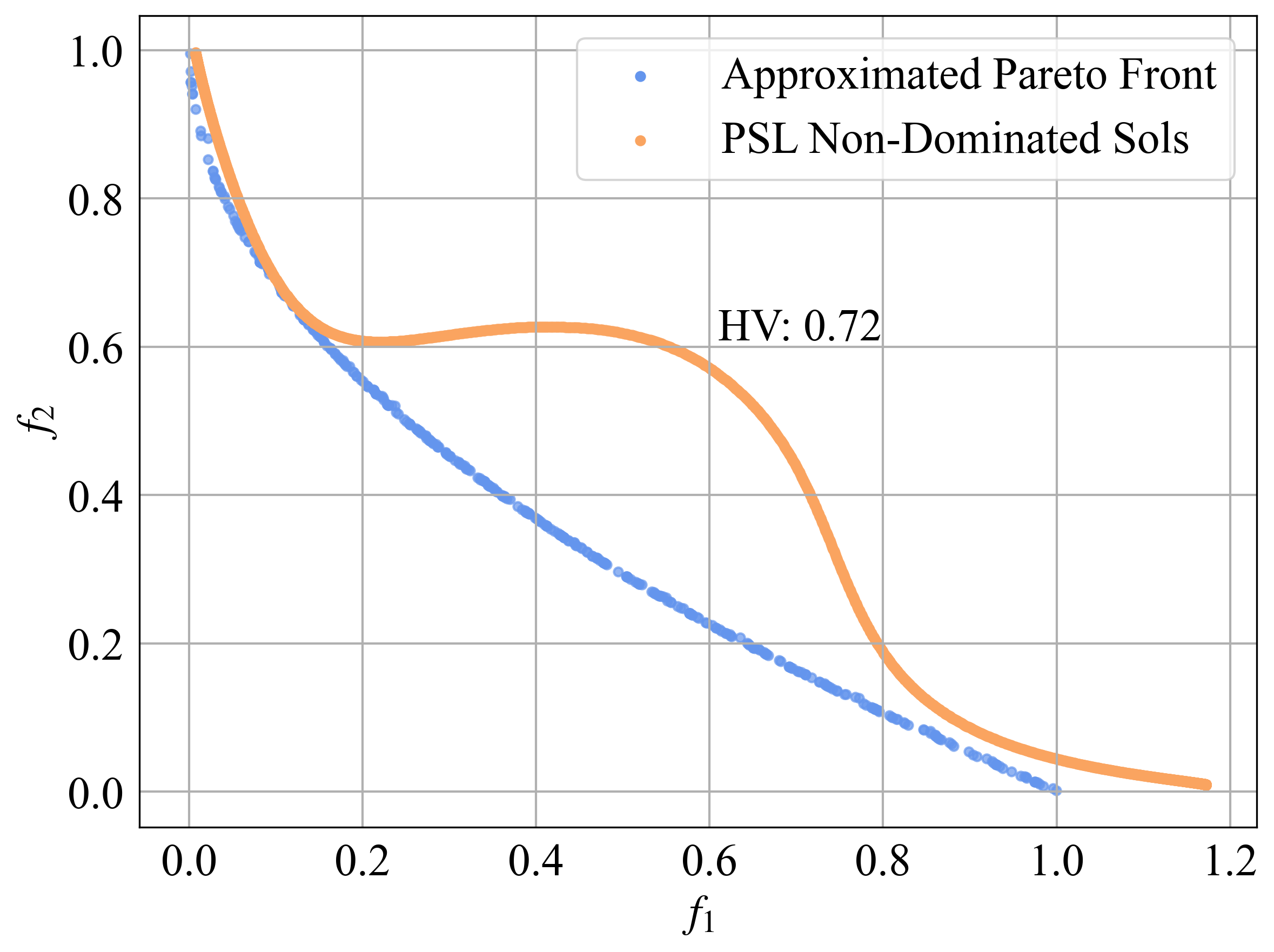}
    \end{minipage}
    }
    \subfigure[CoPSL-LS]{
    \begin{minipage}[b]{0.23\linewidth}
    \centering
    \includegraphics[scale=0.21]{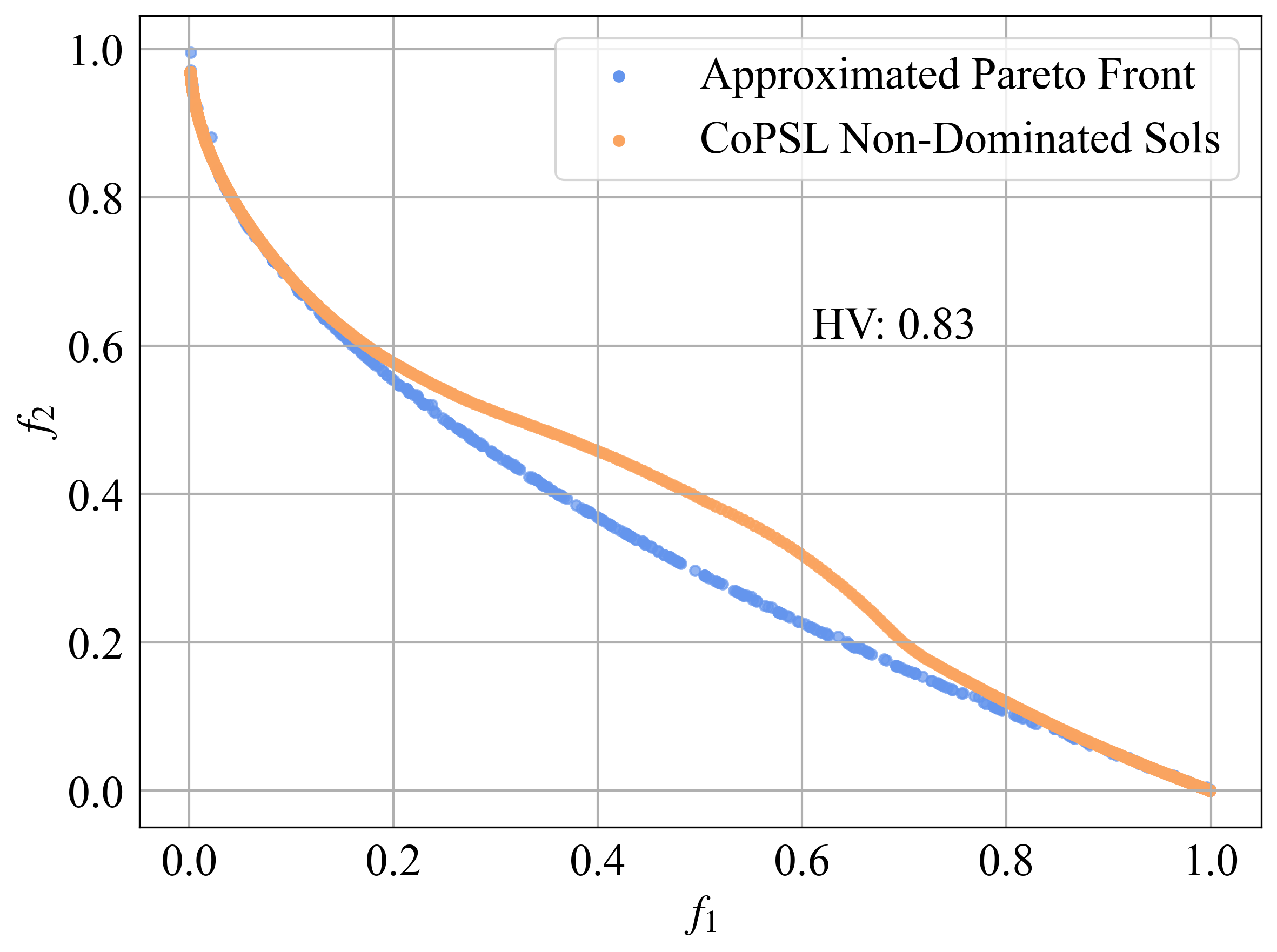}
    \end{minipage}
    }
    \subfigure[CoPSL-COSMOS]{
    \label{copslcosmos2}
    \begin{minipage}[b]{0.23\linewidth}
    \centering
    \includegraphics[scale=0.21]{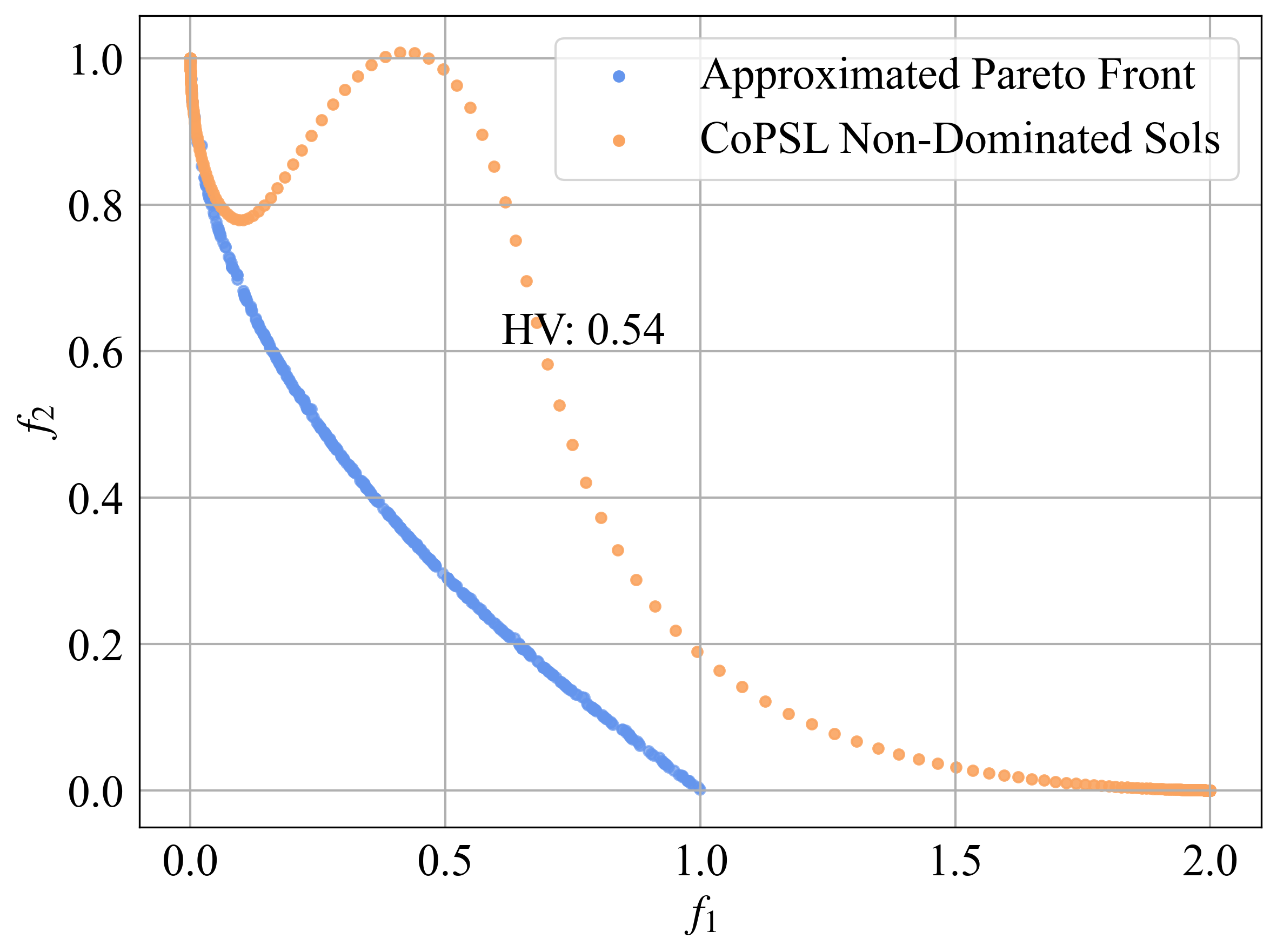}
    \end{minipage}
    }
    \subfigure[CoPSL-TCH]{
    \begin{minipage}[b]{0.23\linewidth}
    \centering
    \includegraphics[scale=0.21]{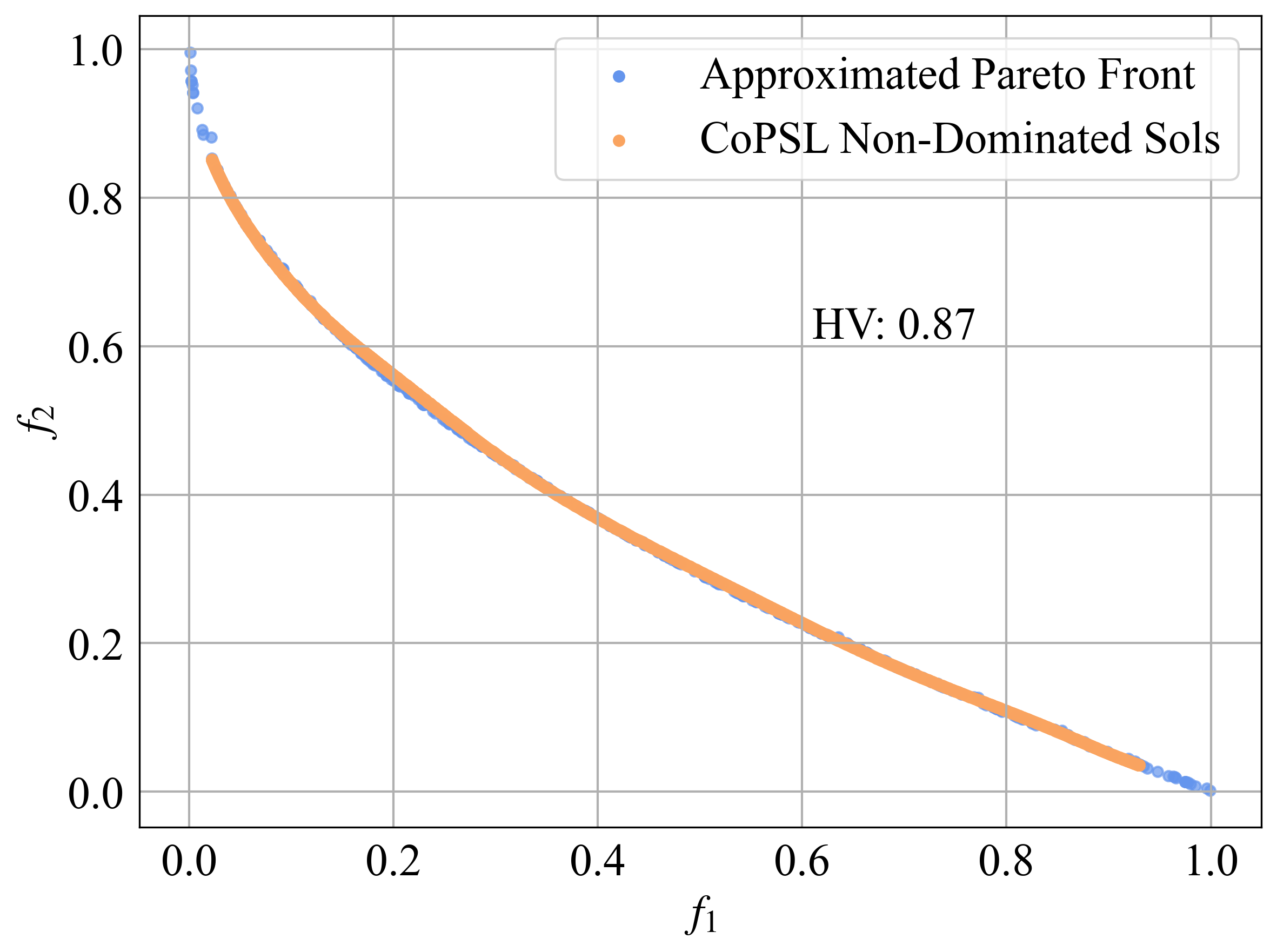}
    \end{minipage}
    }
    \subfigure[CoPSL-MTCH]{
    \label{copslmtch2}
    \begin{minipage}[b]{0.23\linewidth}
    \centering
    \includegraphics[scale=0.21]{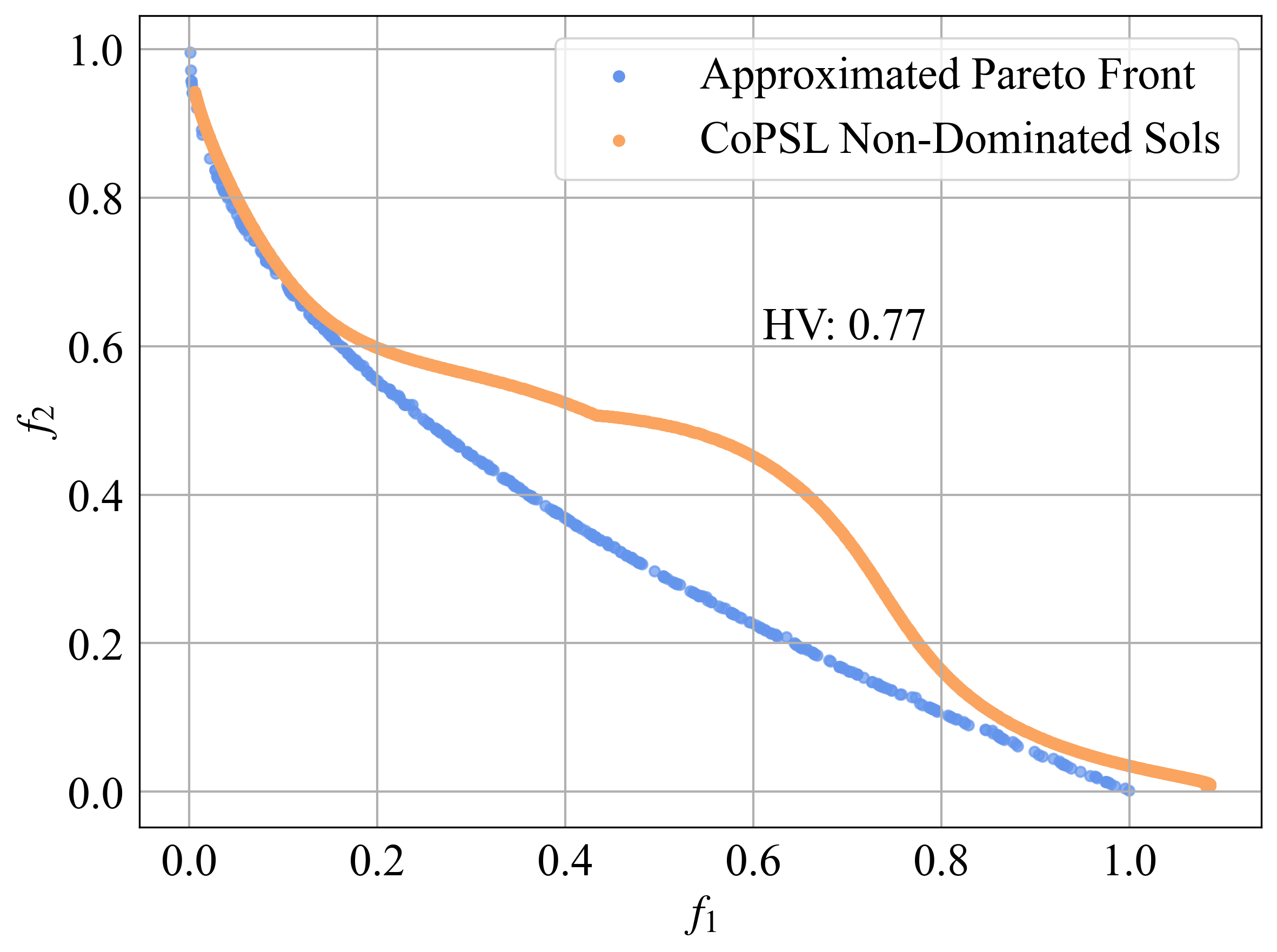}
    \end{minipage}
    }
    \caption{Pareto front comparisons on synthetic benchmark F1. The top part represents PSL, and the bottom part represents CoPSL. Each set of top and bottom comparisons is derived from the same random seed.} \label{fig:exp_pf_apd}
\end{figure*}

To further explore the effectiveness of learning from multiple Pareto sets concurrently in a collaborative manner, we executed a thorough comparative analysis of PSL and CoPSL across multiple synthetic benchmarks and real-world application problems. These comparisons can be viewed as ablation studies, where each CoPSL alternative was benchmarked against a corresponding PSL setup paired with the same loss function. For consistency, we standardized the batch size $B=30$. The results, encapsulated in Tables~\ref{tab:exp_PSL_F} and~\ref{tab:exp_PSL_RE}, showcase the HV indicator and runtime for synthetic and real-world problem sets, respectively. Furthermore, Fig.~\ref{fig:exp_psl} presents the log HV difference to the approximate Pareto front for a subset of the problems during the optimization process.

The results indicate that CoPSL significantly outperforms PSL in terms of efficiency. Employing identical loss functions, CoPSL generally exhibits a marginal but consistent advantage over PSL in both synthetic and real-world problem suites with respect to their approximation capabilities. Specifically, within each problem suite, CoPSL surpasses its PSL counterpart in all but one instance or, in some cases, across the board. Moreover, as depicted in Fig.~\ref{fig:exp_psl}, CoPSL demonstrates more robust updates compared to PSL. PSL, particularly PSL-MTCH as shown in Fig.~\ref{fig:exp_psl_re32}, are susceptible to fluctuations, whereas CoPSL maintains a steadier progression. This suggests that CoPSL is capable of learning more potent shared representations among MOPs, thus enhancing its ability to approximate the Pareto set and improving its generalizability.

Additionally, Fig.~\ref{fig:exp_pf_apd} displays the non-dominated solutions generated by PSL and CoPSL on benchmark F1. The lower section of the figure represents the CoPSL alternatives, while the upper section represents the PSL counterparts. Each top- and bottom-set comparison originates from the same random seed. The results presented in Fig.~\ref{fig:exp_pf_apd} demonstrate that CoPSL is significantly better than PSL, as the non-dominated solutions of CoPSL are much closer to the Pareto set and exhibit a superior distribution. It highlights the advantages of learning across multiple MOPs and the benefits of shared representations. More experimental results are provided in the Appendix.

It should be highlighted that there appears to be little correlation between the real-world problems in the nature of questions and the shape of the Pareto sets. Nonetheless, CoPSL is still capable of concurrently learning an entire suite of MOPs while also learning shared common representations that enhance overall performance. This underscores the remarkable potential of CoPSL to process numerous MOPs simultaneously and still derive valuable shared representations.

\section{Conclusion and Future Work} \label{section6}
In this paper, we proposed the CoPSL framework, a collaborative approach designed to tackle multiple MOPs simultaneously. Inspired by MTL, the framework incorporates an architecture with both shared and MOP-specific layers. The shared layers focus on uncovering commonalities among various MOPs, while the MOP-specific layers are tailored to learn the Pareto set for each MOP. We experimentally revealed that the shared layers excel at capturing shared common representations of preference vectors across different MOPs. This ability facilitates a more generalized representation space, ultimately improving the approximation of the Pareto sets. We compared CoPSL with state-of-the-art EMO and PSL algorithms on a variety of both synthetic and real-world MOPs. Extensive experiments demonstrated CoPSL's superior efficiency and robustness in approximating Pareto sets.

Furthermore, we emphasized the dynamic adjustability of the weight vector in the CoPSL framework during the optimization process. Drawing inspiration from indicator-based EMO algorithms, we suggest leveraging additional indicators in multi-objective optimization to accurately reflect the current state of the solution sets. By dynamically adjusting the weight vector through the introduction of evaluation indicators, we anticipate an enhanced capability to handle multiple MOPs more effectively in CoPSL. This direction holds promise for optimizing the CoPSL framework's performance in solving multiple complex MOPs concurrently, and we believe it merits further exploration in future research.

\bibliographystyle{ieeetr}
\bibliography{refs}

\appendix

\section*{Benchmarks and Real-World Problems}

\begin{table}[!htbp]
\centering
\caption{Problem information and reference point for both synthetic benchmarks and real-world engineering design problems.}
\label{tab:problem}
\begin{tabular}{lccc}
\toprule
Problem                        & $m$ & $n$ & Reference Point       \\
\midrule
F1                             & 2   & 6   & (1.1,1.1)             \\
F2                             & 2   & 6   & (1.1,1.1)             \\
F3                             & 2   & 6   & (1.1,1.1)             \\
F4                             & 2   & 6   & (1.1,1.1)             \\
F5                             & 2   & 6   & (1.1,1.1)             \\
F6                             & 2   & 6   & (1.1,1.1)             \\
Two Bar Truss Design           & 3   & 3   & (550,9.9e+6,2.2e+7)   \\
Welded Beam Design             & 3   & 4   & (38.83,1.9e+4,4.6e+8) \\
Disc Brake Design              & 3   & 4   & (5.83,3.43,27.5)      \\
Vehicle Crashworthiness Design & 3   & 5   & (1865,12.98,0.32)     \\
Rocket Injector Design         & 3   & 4   & (1.08,1.05,1.08)      \\
\bottomrule
\end{tabular}
\end{table}

\begin{figure*}[!t]
    \centering
    \subfigure[PSL-LS]{
    \begin{minipage}[b]{0.23\linewidth}
    \centering
    \includegraphics[scale=0.31]{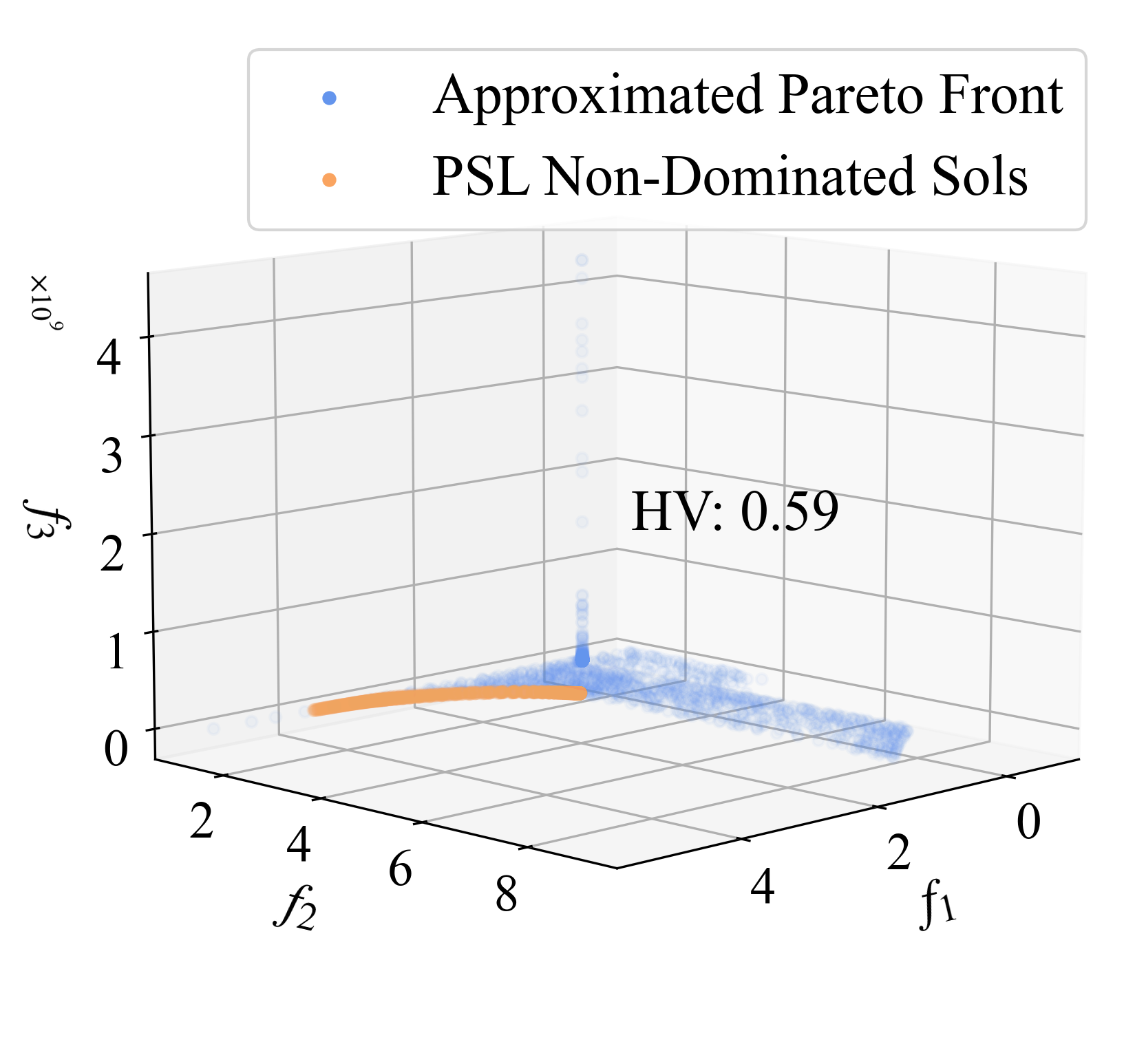}
    \end{minipage}
    }
    \subfigure[PSL-COSMOS]{
    \label{pslcosmos}
    \begin{minipage}[b]{0.23\linewidth}
    \centering
    \includegraphics[scale=0.31]{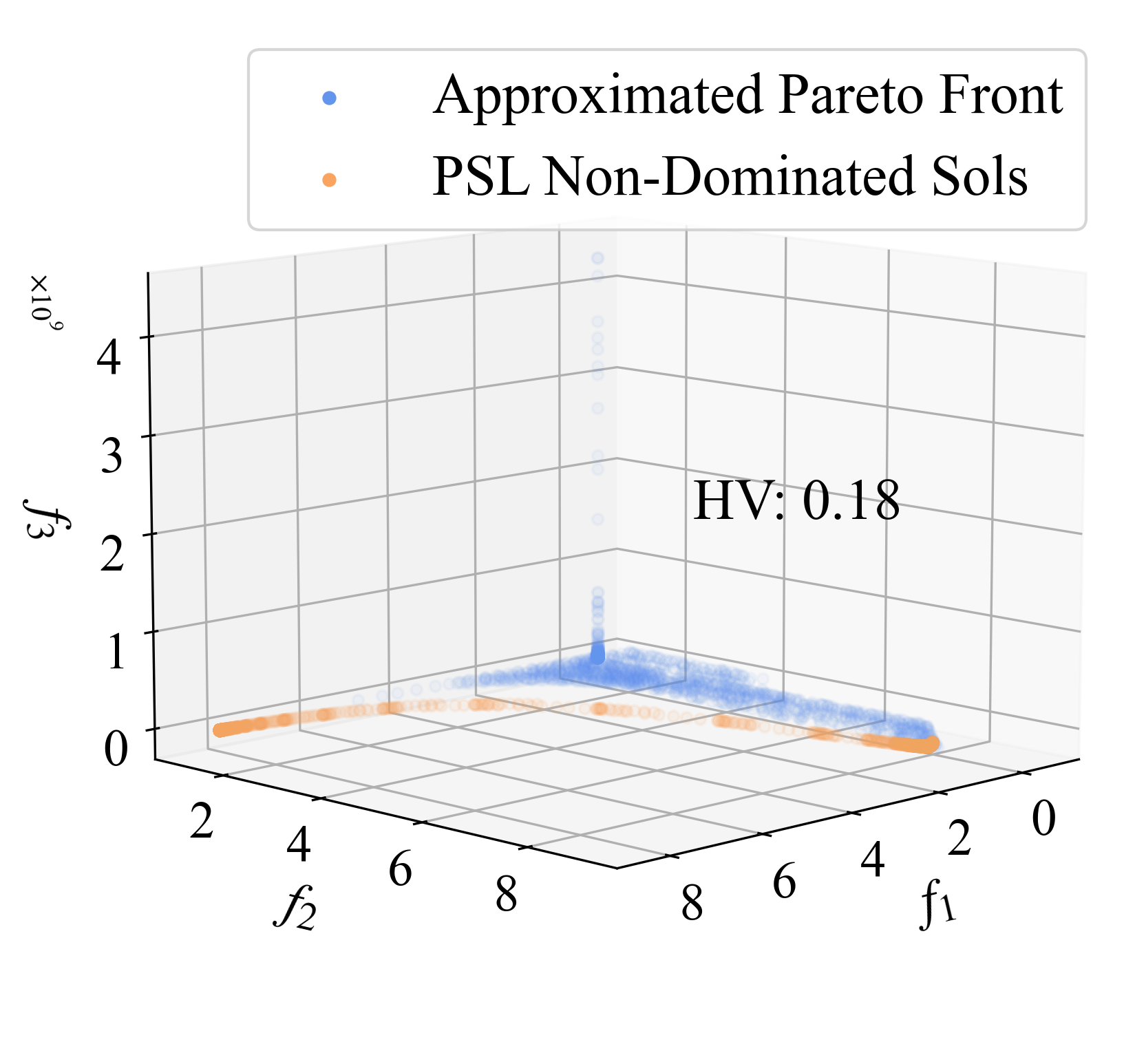}
    \end{minipage}
    }
    \subfigure[PSL-TCH]{
    \begin{minipage}[b]{0.23\linewidth}
    \centering
    \includegraphics[scale=0.31]{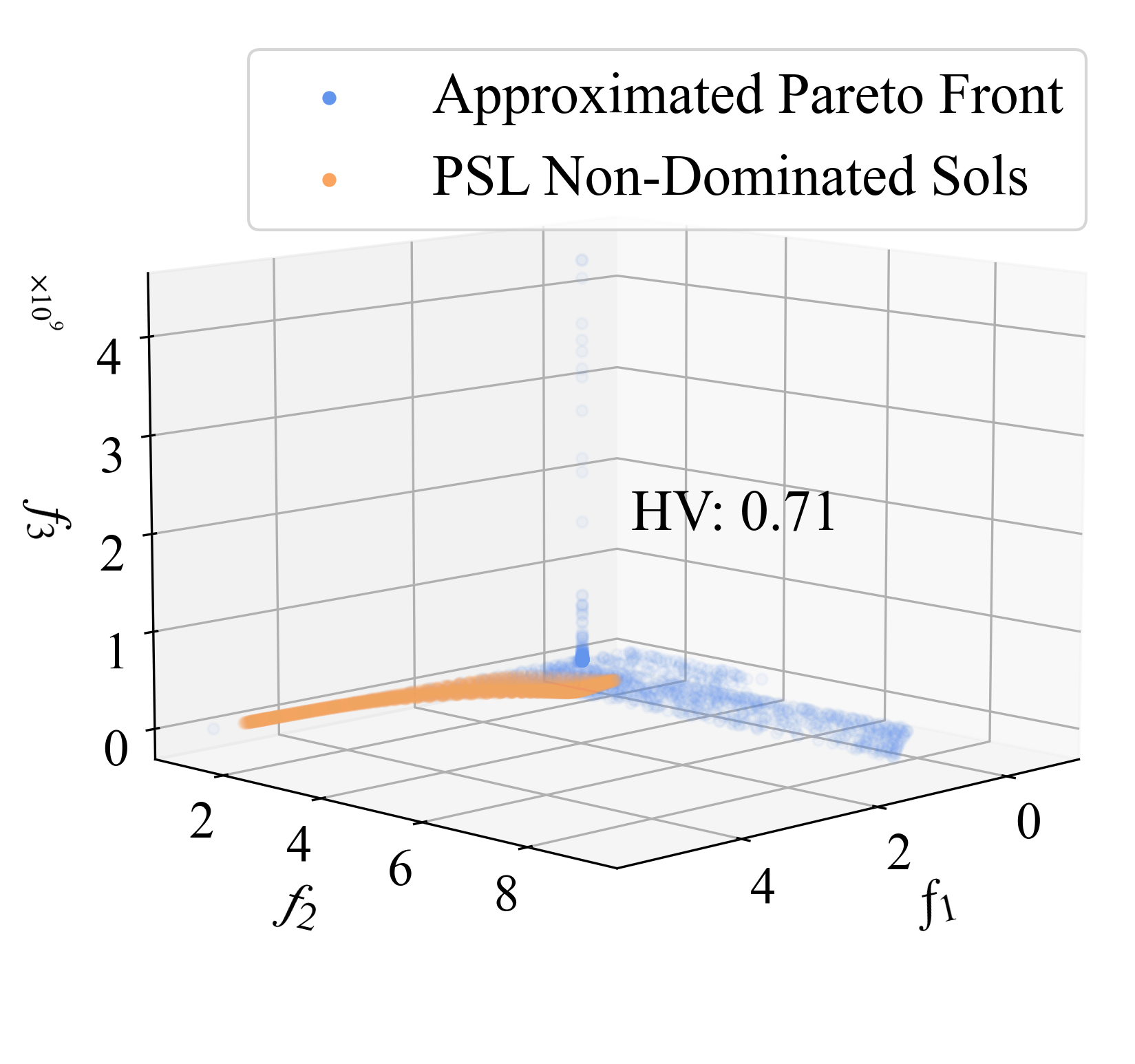}
    \end{minipage}
    }
    \subfigure[PSL-MTCH]{
    \label{pslmtch}
    \begin{minipage}[b]{0.23\linewidth}
    \centering
    \includegraphics[scale=0.31]{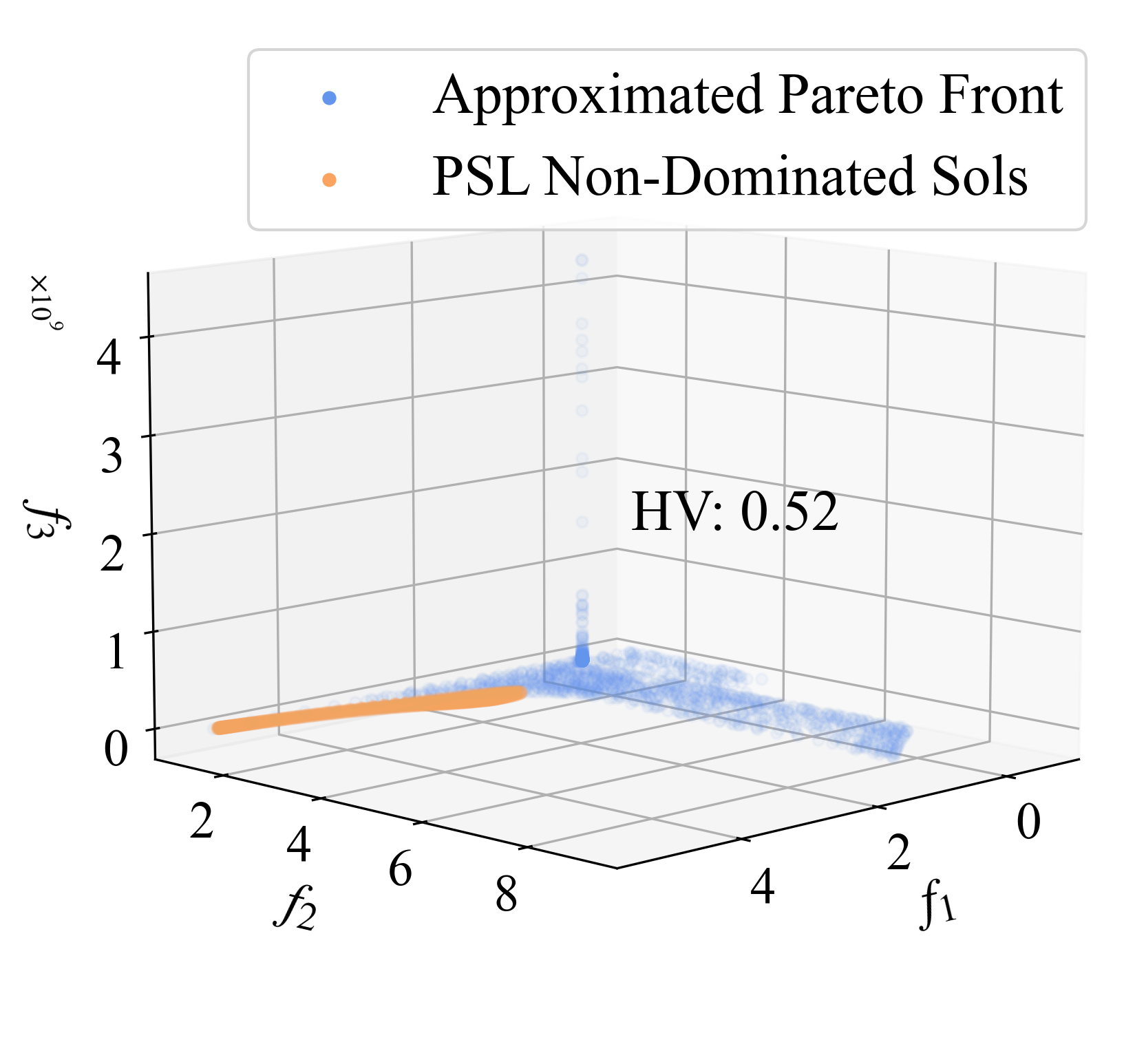}
    \end{minipage}
    }
    \subfigure[CoPSL-LS]{
    \begin{minipage}[b]{0.23\linewidth}
    \centering
    \includegraphics[scale=0.31]{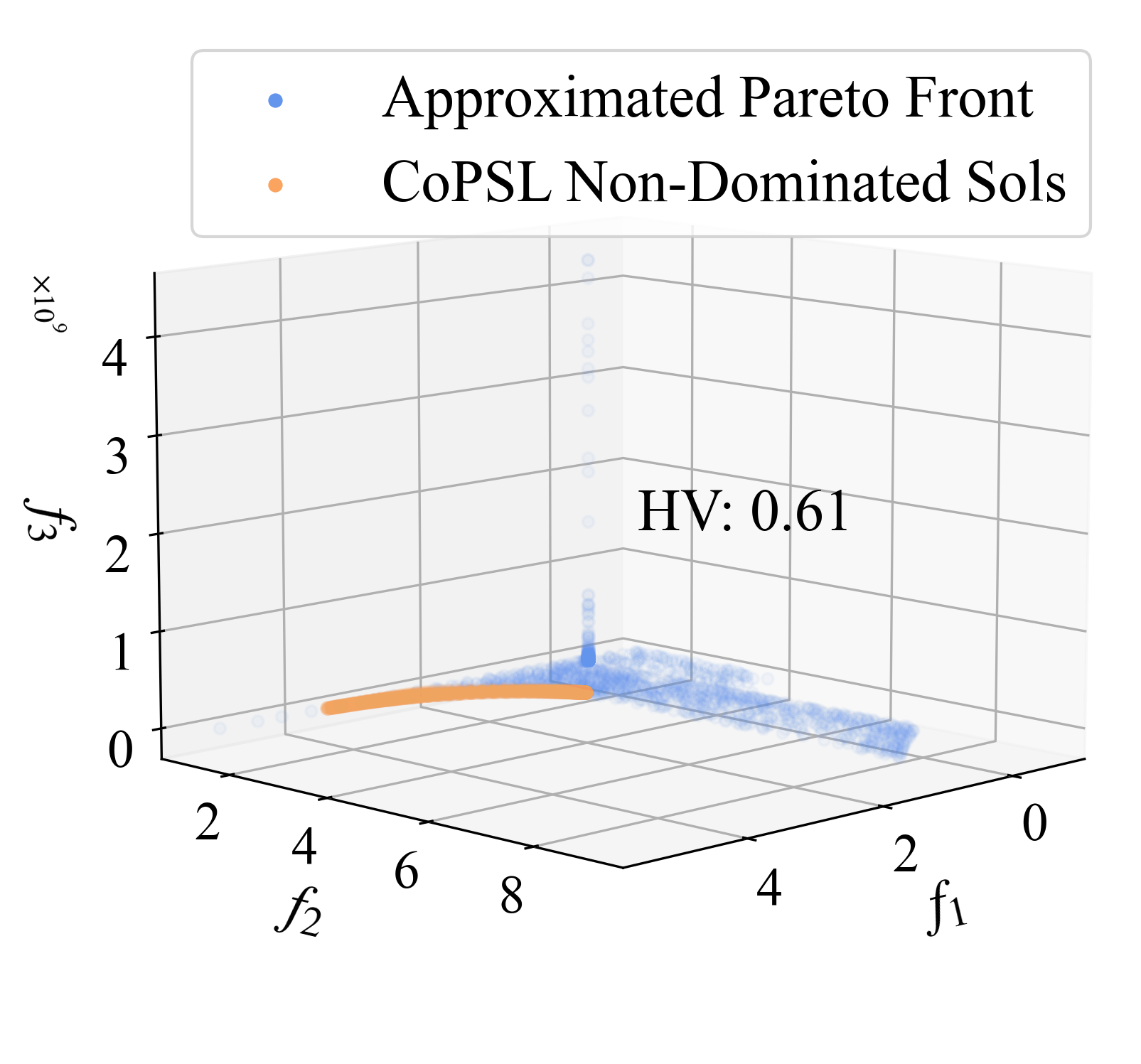}
    \end{minipage}
    }
    \subfigure[CoPSL-COSMOS]{
    \label{copslcosmos}
    \begin{minipage}[b]{0.23\linewidth}
    \centering
    \includegraphics[scale=0.31]{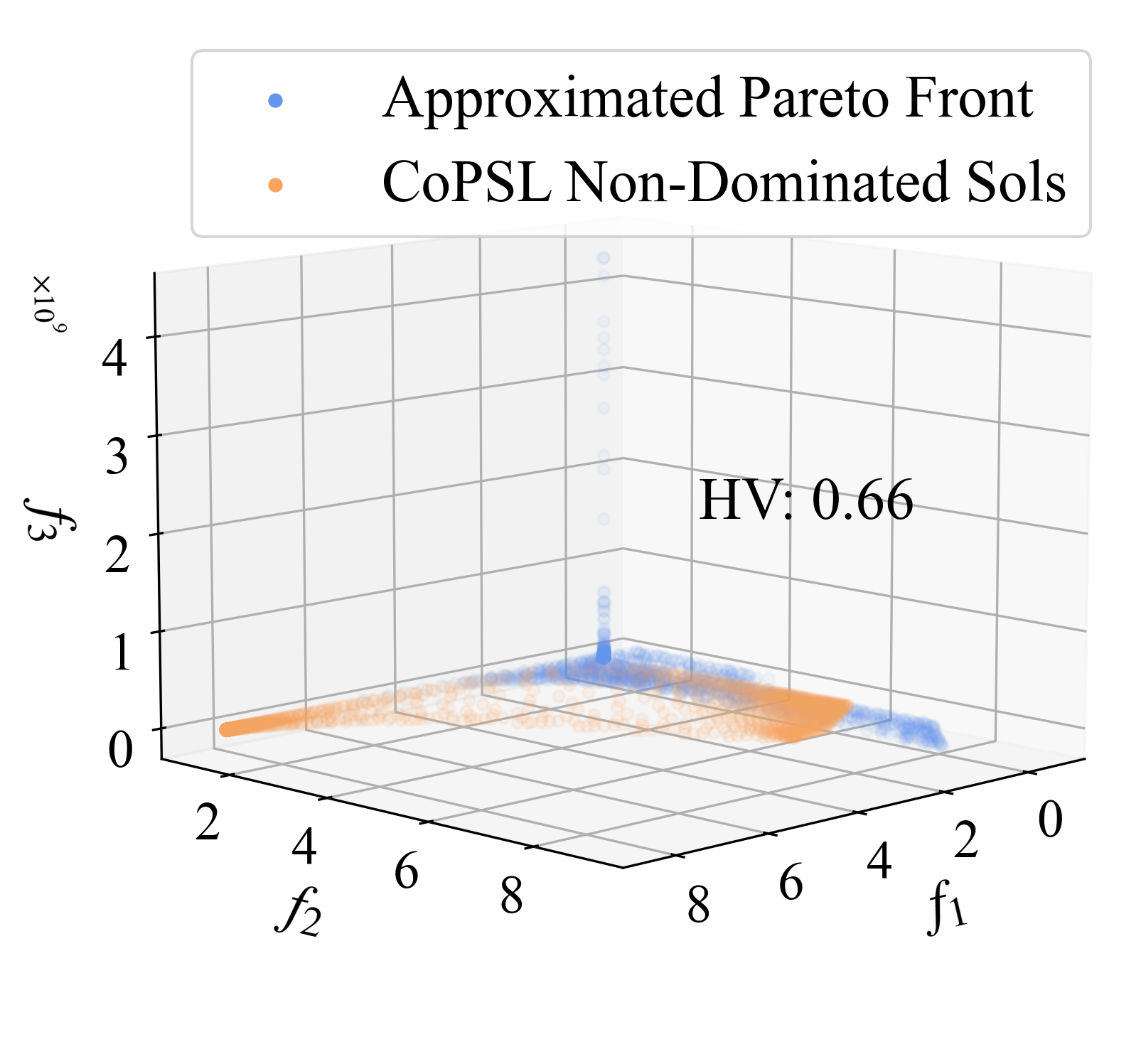}
    \end{minipage}
    }
    \subfigure[CoPSL-TCH]{
    \begin{minipage}[b]{0.23\linewidth}
    \centering
    \includegraphics[scale=0.31]{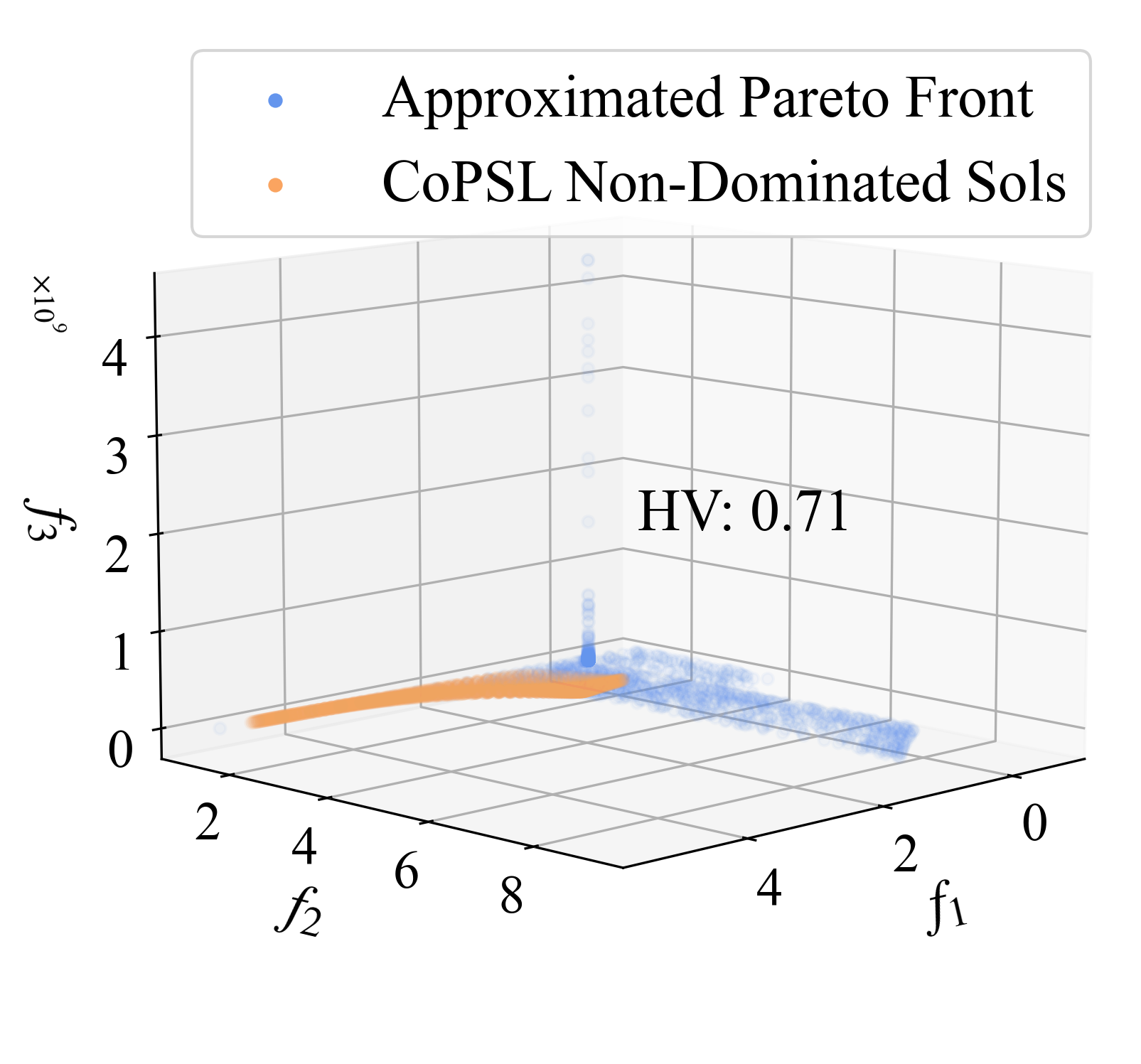}
    \end{minipage}
    }
    \subfigure[CoPSL-MTCH]{
    \label{copslmtch}
    \begin{minipage}[b]{0.23\linewidth}
    \centering
    \includegraphics[scale=0.31]{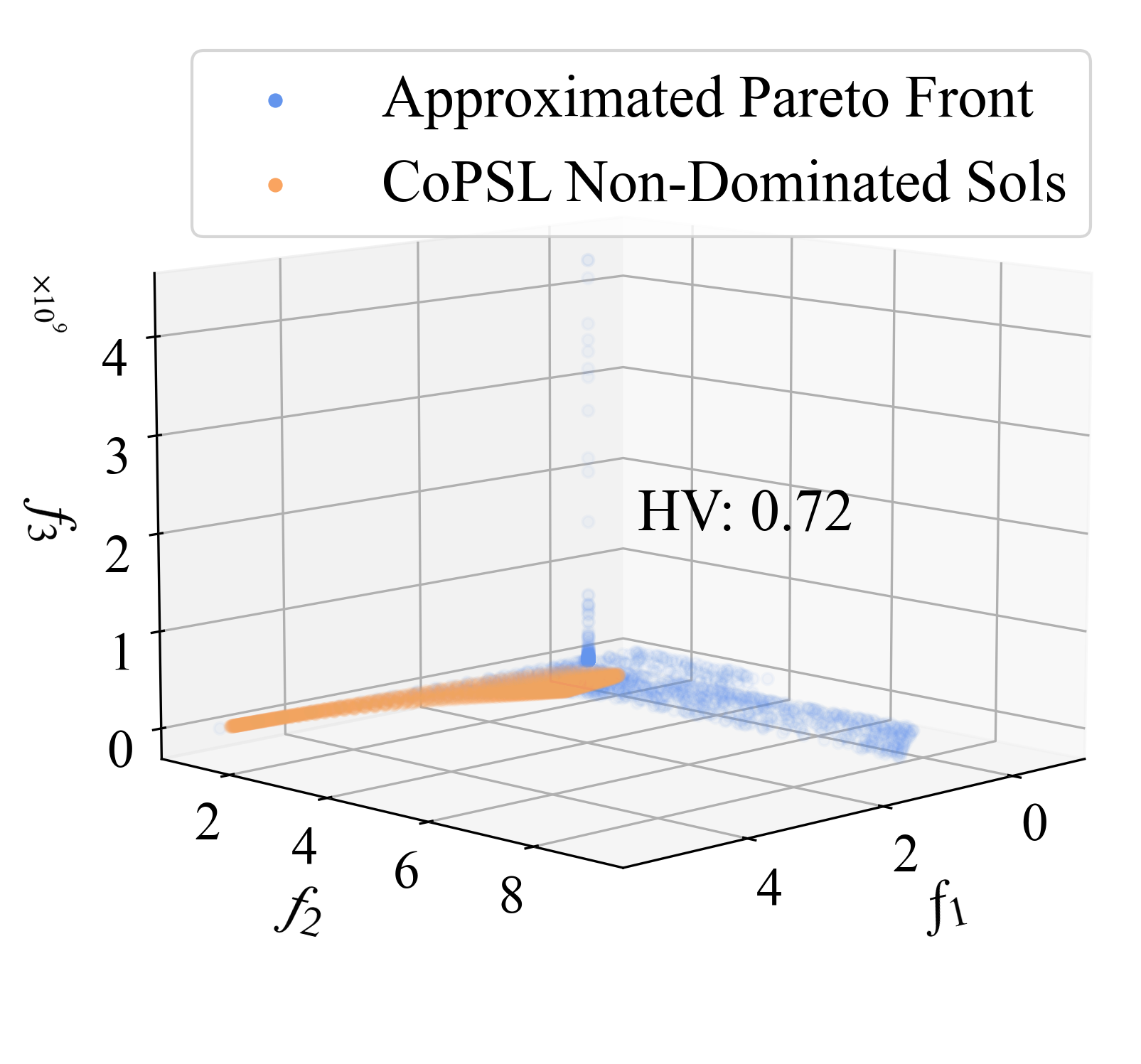}
    \end{minipage}
    }
    \caption{Pareto front comparisons on the disc brake design problem (i.e., RE33). The top part represents PSL, and the bottom part represents CoPSL. Each set of top and bottom comparisons is derived from the same random seed.} \label{fig:exp_pf}
\end{figure*}

To better evaluate our proposed CoPSL approach, we conducted experiments on six synthetic benchmarks~\cite{lin2022pareto} and five real-world multi-objective engineering design problems~\cite{tanabe2020easy}. We utilized the approximated Pareto fronts as the actual fronts to compute the HV values in both suites of problems. The problem information and reference points can be found in Table~\ref{tab:problem}, where the reference points are used to compute the HV indicator.

\section*{Model Structures}

In this work, we employ fully connected networks with varying layers as our experimental neural networks. Notably, we explored different architectural layers in Sections~\ref{section4} and~\ref{section5}.

For our primary experiments detailed in Section~\ref{section5}, we utilized a model with 2 hidden layers, each comprising 256 hidden units, designated as the PSL model. The architecture of this model is outlined as follows:
\begin{align*}
    \text{Input}~\mathbf{p} &\rightarrow \text{Linear}(m,256) \rightarrow \text{ReLU} \\
    &\rightarrow \text{Linear}(256,256) \rightarrow \text{ReLU} \\
    &\rightarrow \text{Linear}(256,n) \rightarrow \text{Sigmoid} \rightarrow \text{Output}~\mathbf{x}(\mathbf{p}) 
\end{align*}

To ensure a fair comparison, we employed a model in CoPSL that is almost equivalent to this PSL model. The CoPSL model comprises one shared layer and two MOP-specific layers, described as follows:
\begin{alignat*}{2}
    \text{Shared layer:}&& &\text{Input}~\mathbf{p} \rightarrow \text{Linear}(m,256) \rightarrow \text{ReLU} \\
    \text{MOP-specific layers:}&& &\text{Linear}(256,256) \rightarrow \text{ReLU} \\
    &&\rightarrow &\text{Linear}(256,n) \rightarrow \text{Sigmoid} \\
    &&\rightarrow &\text{Output}~\mathbf{x}(\mathbf{p}) 
\end{alignat*}

In the experimental investigations in Section~\ref{section4}, we utilized a model with 3 hidden layers, each containing 180 hidden units, as the PSL model. The structure of this model is depicted as follows:
\begin{align*}
    \text{Input}~\mathbf{p} &\rightarrow \text{Linear}(m,180) \rightarrow \text{ReLU} \\
    &\rightarrow \text{Linear}(180,180) \rightarrow \text{ReLU} \\
    &\rightarrow \text{Linear}(180,180) \rightarrow \text{ReLU} \\
    &\rightarrow \text{Linear}(180,n) \rightarrow \text{Sigmoid} \rightarrow \text{Output}~\mathbf{x}(\mathbf{p}) 
\end{align*}
We utilized a CoPSL model nearly equivalent to this PSL model as well. Given that all possible enumerations of the CoPSL architecture have been detailed in Fig.~\ref{fig:UnderstandingCoPSL}, we will not repeat them here.

\section*{Additional Experimental Results}
For a more comprehensive comparison, we provide the non-dominated solutions generated by PSL and CoPSL on the disc brake design problem (i.e., RE33) in Fig.~\ref{fig:exp_pf}. The lower part of the figure represents the CoPSL alternatives, while the upper part represents the PSL counterparts. Each top- and bottom-set comparison originates from the same random seed. Compared with PSL-COSMOS and CoPSL-COSMOS (i.e., Subfig.~\ref{pslcosmos} and~\ref{copslcosmos}), as well as PSL-MTCH and CoPSL-MTCH (i.e., Subfig.~\ref{pslmtch} and~\ref{copslmtch}), the non-dominated solutions of CoPSL are significantly closer to the Pareto set and have better diversity. These experimental results emphasize the advantages of robust shared representations that are learned in a collaborative manner.

\end{document}